\documentclass{article} 
\usepackage{iclr2027_conference,times}

\usepackage{amsmath,amsfonts,bm}

\def\eqref#1{equation~\ref{#1}}

\def\1{\bm{1}}

\DeclareMathAlphabet{\mathsfit}{\encodingdefault}{\sfdefault}{m}{sl}
\SetMathAlphabet{\mathsfit}{bold}{\encodingdefault}{\sfdefault}{bx}{n}

\usepackage{hyperref}
\usepackage{url}

\usepackage{booktabs}
\usepackage[table]{xcolor}   
\usepackage{graphicx}        
\usepackage[most]{tcolorbox}
\usepackage{graphicx}

\usepackage{amsthm}

\newtheorem{proposition}{Proposition}
\newtheorem{corollary}[proposition]{Corollary}
\usepackage{natbib}
\usepackage{caption}
\usepackage{wrapfig}
\usepackage{tabularx}
\usepackage{array}
\usepackage{arydshln}
\usepackage{tikz}
\usepackage[most]{tcolorbox}
\usepackage{amssymb}

\usepackage{algorithm}     
\usepackage{algorithmic}   
\usepackage{xcolor}        
\usepackage{amsmath}       

\newcommand{\sd}[1]{\raisebox{-0.3ex}{\scalebox{0.63}{$(#1)$}}}
\newcommand{\tr}[1]{#1}

\title{Cross-Rollout Bellman Closure for Long-Horizon Agentic Reinforcement Learning}

\author{%
  \textbf{Yangyang Ren}$^{1,2}$\thanks{Equal contribution.} \quad
  \textbf{Haodong Zhu}$^{1,2}$\footnotemark[1] \quad
  \textbf{Linlin Yang}$^{3}$\thanks{Corresponding author: Linlin Yang (\texttt{lyang@cuc.edu.cn}).} \quad
  \textbf{Sheng Xu}$^{3}$ \quad
  \textbf{Peichao Lai}$^{4}$ \\
  \textbf{Baochang Zhang}$^{1,5}$ \\[0.5em]
  $^{1}$Beihang University \quad
  $^{2}$Zhongguancun Academy \quad
  $^{3}$Communication University of China \\
  $^{4}$Peking University \quad
  $^{5}$Hangzhou Innovation Institute of Beihang University
}

\iclrfinalcopy 
\begin{document}

\maketitle
\lhead{}

\begin{abstract}
Group-based reinforcement learning such as GRPO has become a standard recipe for post-training LLM agents, replacing a learned critic with relative comparison among rollouts sampled for each task. 
\tr{In long-horizon settings, these rollouts revisit shared anchor states, offering cross-rollout evidence for step-level credit. }
%
Ideally, step-level credit should incorporate evidence beyond the realized suffixes observed at an anchor while aggregating alternative continuations according to their empirical frequencies.
Existing estimators capture only one of these properties: visit-local averaging pools realized suffix returns at shared anchors and respects observed frequencies, but does not recursively propagate evidence across rollouts;
whereas shortest-path estimators have global reach but allow a rarely observed route to dominate an anchor's value.
%
Instead, we introduce a \textbf{C}ross-\textbf{R}ollout \textbf{B}ellman \textbf{C}losure (\textbf{CRBC}) method, which merges each rollout group into a finite empirical process with absorbing success and failure boundaries, and evaluates its behaviour-policy Bellman fixed point with one linear solve. 
\tr{This fixed point uses the same empirical action and transition frequencies to propagate evidence through shared anchors and aggregate alternative continuations.}
Backing up the resulting state values through observed transitions yields action values, whose gain over the corresponding state value provides step-level credit.
A corresponding finite-depth family recovers visit-local return averaging at zero depth and converges to the exact closure as depth increases. 
The normalized closure credit is combined with the trajectory-level group advantage for policy optimization, without additional environment rollouts. 
Across ALFWorld, WebShop, and Sokoban benchmarks with multiple model scales, CRBC consistently improves final performance and learning efficiency. 
For example, CRBC outperforms the strongest baseline by 5.59 percentage points on ALFWorld with Qwen2.5-1.5B-Instruct, achieving a new state-of-the-art performance.
\end{abstract}
\section{Introduction}
\label{sec:introduction}
Large Language Models (LLMs) \citep{achiam2023gpt,yang2024qwen2,guo2025deepseek} increasingly serve as interactive agents in embodied worlds, web interfaces, and tool-augmented workflows, where task completion requires planning across multiple interdependent decisions \citep{shridhar2021alfworld,yao2022webshop,furuta2024multimodal,schick2023toolformer}. Reinforcement learning (RL) provides a post-training approach to improving these capabilities \citep{ziegler2019fine,ouyang2022training,team2025kimi}. Group-based methods such as RLOO and GRPO replace PPO's learned critic \citep{schulman2017proximal} with relative comparisons among same-task rollouts, avoiding the cost of training a separate value network \citep{ahmadian2024basicsrevisitingreinforcestyle,deepseek-math,yu2025dapo}.

\begin{figure*}[t]
\centering
\includegraphics[width=\textwidth]{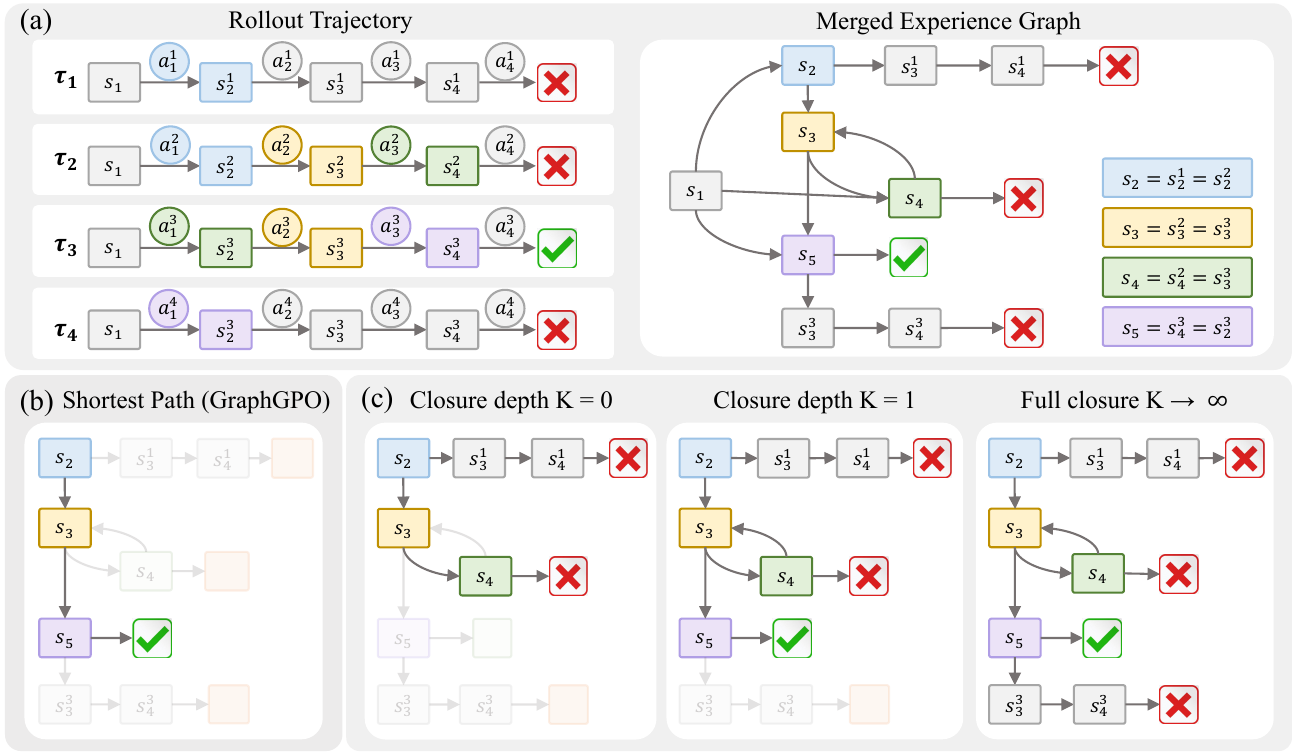}
\caption{Reach and aggregation for step-level credit. \tr{For the rollout trajectory $\tau$, node $s$ denotes an \emph{anchor}: an environment-relevant state representation that may be shared by multiple visits, and the edge represents an observed transition after executing semantic action $a$. } (a) Shared anchors merge rollouts into a finite empirical process. (b) Shortest-path estimation aggregates by an extremum, ignoring transition frequencies; a rarely executed continuation can dominate anchor value. (c) The finite-depth family on this process, illustrated at $s_2$, uses behaviour-consistent expectations: $K=0$ pools visit-local suffix returns; $K=1$ adds one cross-rollout backup; and $K\to\infty$ evaluates all supported continuations at the behaviour-policy Bellman fixed point (CRBC), giving global reach.}
\label{fig:motivation}
\vspace{-15pt}
\end{figure*}

A single trajectory-level advantage, when broadcast to every step, cannot distinguish the contributions of individual actions \citep{wang2025ragen,jin2025search,chen2025reinforcement}. Yet rollouts sampled from the same task and initial condition often visit shared intermediate situations, or \emph{anchors}, providing transition evidence beyond terminal outcomes. Step-level methods exploit this structure by comparing visits at shared anchors, with some further refining comparison groups by the consistency of preceding histories \citep{feng2025group,wang2026rtmcsteplevelcreditassignment,he2026hierarchyofgroupspolicyoptimizationlonghorizon}. A further line pools transitions across rollouts into a merged graph and scores actions on it \citep{cheng2026beyond,wang2026groupgraphpolicyoptimizationlonghorizon}. Together, these approaches expose a reusable cross-rollout structure within each task group (Figure~\ref{fig:motivation}(a)).

Yet exposing this structure determines which observed transitions may be recombined, not how they should be evaluated. Two choices remain open: how far evidence may travel (\emph{reach}), and how the evidence that arrives is combined (\emph{aggregation}).
Visit-local averaging \citep{feng2025group} aggregates by an expectation over the observed behaviour, averaging realized suffix returns across visits to an anchor, but its reach is local: each occurrence is scored by its own continuation, 
so evidence pooled at shared successors is not recursively propagated back to the current anchor (Figure~\ref{fig:motivation}(c)).
A shortest-path estimator \citep{cheng2026beyond} has global reach over the merged graph, but aggregates by an extremum (Figure~\ref{fig:motivation}(b)): it values the shortest route to success without accounting for how frequently its transitions were observed, so a rarely observed continuation can dominate the estimate.
%
Intuitively, an action's credit should account for both successful and failed continuations revealed by other rollouts at a shared successor.
These limitations motivate combining global reach with behaviour-consistent aggregation on the group's empirical process,  \tr{ leading to our central question: \emph{once a rollout group has been merged into an empirical process, how should step credit propagate through and aggregate its supported continuations?}}

\begin{figure*}[t]
\centering
\includegraphics[width=\textwidth]{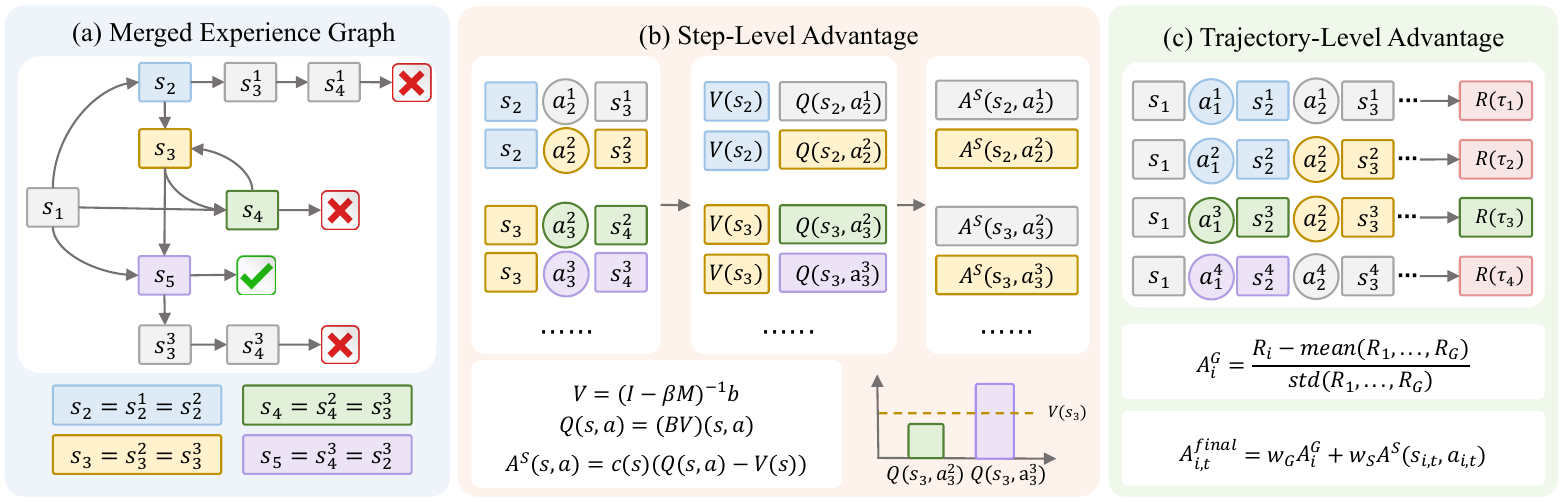}
\caption{\textbf{Overview of CRBC.} (a) Shared anchors merge rollouts into an empirical process with absorbing success and failure boundaries. (b) A Bellman fixed-point solve yields \tr{the state value} $V$, followed by action backups to obtain \tr{the action value} $Q$ and normalized $Q-V$ credit, \tr{which gives relative action credit}. (c) Step-level credit and trajectory-level group-relative advantages are combined for policy optimization.}
\label{fig:framework}
\vspace{-15pt}
\end{figure*}

To address this issue, we introduce \textbf{C}ross-\textbf{R}ollout \textbf{B}ellman \textbf{C}losure (\textbf{CRBC}), which merges each rollout group into a finite empirical process over shared anchors and evaluates it at its behaviour-policy Bellman fixed point with one linear solve (Figure~\ref{fig:framework}).
\tr{Specifically, the Bellman closure repeatedly
propagates successor values backward through the observed transitions until
each anchor value is consistent with the values of its possible continuations.}
This closure is also the limit of Bellman backups initialized with visit-local suffix averages, allowing anchors to incorporate evidence from rollouts that never visited them. 
With success and failure as absorbing boundaries, this fixed point is the state value $V$, the exact discounted probability of reaching success under the group's own behaviour. Backing $V$ up through the observed transitions gives the action value $Q$, and $Q-V$ measures how much an action improves on that behaviour at the same anchor.
On this empirical process, feasible policy reweighting along the closure-credit direction does not decrease discounted success value (Proposition~\ref{prop:alignment}).
We combine the normalized closure credit with the trajectory-level group advantage for critic-free policy optimization, without additional environment rollouts.
%
%
Our main contributions are summarized as follows:

\begin{itemize}

\item We construct a finite empirical process from each rollout group's transition counts, allowing credit at an anchor to use downstream evidence from rollouts that never visited it. This evidence is weighted by observed action and transition frequencies.

\item We introduce CRBC, which computes the group's behaviour-policy Bellman fixed point with one linear solve and scores each action by its gain over the state value, giving step credit with global reach and behaviour-consistent aggregation.

\item We show that feasible policy reweighting along the closure-credit direction does not decrease discounted success value on the fixed empirical process (Proposition~\ref{prop:alignment}). Experiments on ALFWorld, WebShop, and Sokoban demonstrate gains over the evaluated baselines, including a $5.59$-percentage-point improvement over the strongest baseline on ALFWorld with Qwen2.5-1.5B-Instruct.

\end{itemize}

\section{Related Work}
\label{sec:related}

\paragraph{Reinforcement learning for LLMs and agents.}
LLM post-training uses human feedback \citep{ziegler2019fine,stiennon2020learning,ouyang2022training}, preference optimization \citep{rafailov2024direct}, and verifiable rewards \citep{guo2025deepseek,team2025kimi}. Group-based methods replace PPO's critic \citep{schulman2017proximal} with comparisons among same-prompt responses, including RLOO \citep{ahmadian2024basicsrevisitingreinforcestyle}, GRPO \citep{deepseek-math}, Dr.\ GRPO, DAPO, and CPPO \citep{liu2025understanding,yu2025dapo,lin2025cppo}. Agent training progressed from DQN in text games \citep{mnih2015human,narasimhan2015language} to PPO- and AWR-style updates \citep{peng2019advantage} for embodied and device-control tasks \citep{zhai2024fine,bai2024digirl}. ArCHer \citep{zhou2024archer} learns hierarchical values for tasks such as WebShop \citep{yao2022webshop}, while Agent Q \citep{putta2024agent} uses Monte Carlo tree search \citep{silver2017mastering}. LOOP \citep{chen2025reinforcement} combines leave-one-out estimation with PPO-style updates on AppWorld \citep{trivedi2024appworld}, and RAGEN \citep{wang2025ragen} studies multi-turn trajectory optimization. Sparse rewards in long-horizon tasks such as ALFWorld \citep{shridhar2021alfworld} motivate finer credit assignment.

\paragraph{Step-level credit assignment in agentic RL.}
Step-level credit can use process reward models \citep{ICLR2024_aca97732}, LLM judges \citep{wang2026enhancingllmbasedsearchagents}, learned action-boundary critics \citep{wang2026capocriticguidedactionalignedpolicy}, or rollout-group statistics. GiGPO \citep{feng2025group} compares repeated anchor visits, HGPO \citep{he2026hierarchyofgroupspolicyoptimizationlonghorizon} refines groups by history consistency, and RTMC \citep{wang2026rtmcsteplevelcreditassignment} averages returns over rollout trees. GraphGPO \citep{cheng2026beyond} scores merged graphs by shortest-path distance to success, while G2PO \citep{wang2026groupgraphpolicyoptimizationlonghorizon} derives edge credit from one-step TD errors. Other methods calibrate credit under limited rollout evidence \citep{li2026densercreditenoughevidencecalibrated}, add progress signals for all-failing groups \citep{yang2026progressconditionedgrouppolicyoptimization}, derive credit from state potentials \citep{fan2026progressreliabilityorientedgrouppolicy}, or extract attribution from model reasoning \citep{tan2026hindsightcreditassignmentlonghorizon}. Unlike return averaging, shortest-path scoring, or one-step TD credit, CRBC explicitly solves for the exact behavior-policy Bellman fixed point of the group's empirical process, using observed action and transition frequencies for cross-rollout propagation and aggregation.

\section{Preliminaries}
\label{sec:prelim}

\paragraph{Rollout groups and anchors.}
We consider an LLM agent with policy $\pi_\theta$, parameterized by $\theta$, solving a task $g$ through $G$ rollouts from the same initial condition. At step $t$ of rollout $i$, it observes a language context $h_{i,t}$, generates a response $y_{i,t}\sim\pi_\theta(\cdot\mid h_{i,t})$, and executes the semantic action $a_{i,t}=\mathrm{Exec}(y_{i,t})$, after which the environment returns $h_{i,t+1}$. Rollout $i$ ends after $T_i\le T_{\max}$ steps with a binary success indicator $R_i\in\{0,1\}$ and no intermediate reward. An environment-aware map $\phi:\mathcal{H}\to\mathcal{S}$ from language contexts to anchors defines $s_{i,t}=\phi(h_{i,t})$, with anchors matched only within each task group \citep{feng2025group,cheng2026beyond}. We represent the group by
\begin{equation}
\mathcal{D}_g=\{\tau_i\}_{i=1}^{G},
\qquad
\tau_i=\bigl((s_{i,t},a_{i,t},x_{i,t})_{t=1}^{T_i},\,R_i\bigr),
\label{eq:rollout-group}
\end{equation}
where $x_{i,t}=s_{i,t+1}$ for $t<T_i$, and the terminal successor $x_{i,T_i}$ is the absorbing success boundary $z^{+}$ if $R_i=1$, or the absorbing failure boundary $z^{-}$ otherwise. We write $\mathcal{S}_g$ for the observed non-terminal anchors, $\mathcal{A}_g(s)$ for the actions observed at $s$, and $\mathcal{X}_g=\mathcal{S}_g\cup\{z^{+},z^{-}\}$. Contexts and responses are retained for policy optimization. Time-limit truncations are treated as failures in this representation.

\paragraph{Group-based RL.}
Let $A_{i,t}$ denote the advantage assigned to environment step $(i,t)$. The optimizer maximizes a clipped surrogate with a reference-policy penalty,
\begin{equation}
\mathcal{J}(\theta)=\mathbb{E}\Bigl[\min\bigl(\rho_{i,t}(\theta)A_{i,t},\;
\mathrm{clip}(\rho_{i,t}(\theta),1\pm\epsilon)A_{i,t}\bigr)\Bigr]
-\lambda_{\mathrm{KL}}\,
\mathbb{D}_{\mathrm{KL}}\!\left[\pi_{\theta}\,\|\,\pi_{\mathrm{ref}}\right].
\label{eq:base-objective}
\end{equation}
Here $\mathbb{E}$ averages uniformly over sampled groups and, within each group, over environment steps. $\epsilon$ sets the clipping range, and $\lambda_{\mathrm{KL}}$ weights the KL penalty against the fixed reference policy $\pi_{\mathrm{ref}}$. The importance ratio is $\rho_{i,t}(\theta)=\pi_\theta(y_{i,t}\mid h_{i,t})/\pi_{\theta_{\mathrm{old}}}(y_{i,t}\mid h_{i,t})$, where $\pi_{\theta_{\mathrm{old}}}$ is the rollout policy. Group-relative methods form the advantage by comparing outcomes within the group rather than against a learned value function; GRPO \citep{deepseek-math} uses
\begin{equation}
A^{\mathrm G}_i=
\frac{R_i-\mathrm{mean}(R_1,\dots,R_G)}
{\mathrm{std}(R_1,\dots,R_G)},
\label{eq:group-advantage}
\end{equation}
and RLOO \citep{ahmadian2024basicsrevisitingreinforcestyle} a leave-one-out variant. Ours is agnostic to the choice, and we denote by $A^{\mathrm G}_i$ whatever the underlying optimizer outputs. Setting $A_{i,t}=A^{\mathrm G}_i$ for every step of $\tau_i$ gives one scalar per trajectory, broadcast uniformly, which cannot distinguish the actions inside it.

\paragraph{Step-level credit.}
Step-level methods exploit shared anchors to replace uniform trajectory-level credit with signals that vary within a trajectory. Let $\mathcal{I}_g(s):=\{(i,t):s_{i,t}=s\}$ collect occurrences of anchor $s$, and let $G_{i,t}=\beta^{T_i-t+1}R_i$ denote the realized discounted return, with discount $\beta\in(0,1)$. The step-level advantage is
\begin{equation}
A^{\mathrm step}_{i,t}
=
\frac{G_{i,t}-\mathrm{mean}\{G_{i',t'}:(i',t')\in\mathcal{I}_g(s_{i,t})\}}
{\mathrm{std}\{G_{i',t'}:(i',t')\in\mathcal{I}_g(s_{i,t})\}},
\label{eq:step-baseline}
\end{equation}
combined with $A^{\mathrm G}_i$ by a weighted sum \citep{he2026hierarchyofgroupspolicyoptimizationlonghorizon,feng2025group}. Here \eqref{eq:step-baseline} scores each occurrence by its own realized suffix return, so evidence from other rollouts at shared successors never propagates back to $s$.

\section{Cross-Rollout Bellman Closure}
\label{sec:method}

\tr{To propagate and aggregate evidence from the observed continuations of
rollouts that share anchors, }this section constructs a finite empirical process from a rollout group (Section~\ref{sec:process}), computes its Bellman closure (Section~\ref{sec:closure}), analyses credit alignment across propagation depths (Section~\ref{sec:family}), and integrates closure credit into policy optimization (Section~\ref{sec:credit}), as summarized in Figure~\ref{fig:framework}. Throughout the section and its proofs, $P_g$, $\mu_g$, and all associated Bellman operators, values, and credits refer to the process induced by group $g$.

\subsection{The empirical process}
\label{sec:process}
\tr{We first turn the rollouts in a group into a common empirical process over
shared anchors. This representation preserves the observed action and
transition frequencies needed to propagate and aggregate cross-rollout
evidence.}
\paragraph{Empirical model.}
Let $N_g(s,a,x)$ count occurrences of transition $(s,a,x)$ in $\mathcal{D}_g$, with marginals $N_g(s,a)=\sum_{x\in\mathcal{X}_g}N_g(s,a,x)$ and $N_g(s)=\sum_{a\in\mathcal{A}_g(s)}N_g(s,a)$. The group induces an empirical transition kernel and an empirical behaviour policy over anchors,
\begin{equation}
P_g(x\mid s,a):=\frac{N_g(s,a,x)}{N_g(s,a)},
\qquad
\mu_g(a\mid s):=\frac{N_g(s,a)}{N_g(s)}.
\label{eq:empirical-model}
\end{equation}
The unsmoothed frequencies in \eqref{eq:empirical-model} define the group-induced process, with zero probability assigned to unobserved transitions. Our analysis concerns this empirical process rather than the unknown environment dynamics.

\paragraph{The visit-local estimator.}
Let $\mathcal{I}_g(s,a):=\{(i,t)\in\mathcal{I}_g(s):a_{i,t}=a\}$ collect occurrences of $(s,a)$ within group $g$, so that $|\mathcal{I}_g(s,a)|=N_g(s,a)$. Using the realized discounted returns $G_{i,t}$ defined in Section~\ref{sec:prelim}, we define the visit-local estimates
\begin{equation}
Q_g^{(0)}(s,a):=
\frac{1}{|\mathcal{I}_g(s,a)|}
\sum_{(i,t)\in\mathcal{I}_g(s,a)}G_{i,t},
\qquad
V_g^{(0)}(s):=
\sum_{a\in\mathcal{A}_g(s)}
\mu_g(a\mid s)\,Q_g^{(0)}(s,a).
\label{eq:v0}
\end{equation}
Weighting these action values by their empirical action frequencies makes $V_g^{(0)}(s)$ equal to the mean return over all visits to $s$. Each occurrence contributes only the discounted outcome of its own complete continuation to termination.

\subsection{The Bellman closure}
\label{sec:closure}
\tr{Given the empirical process, we now evaluate how success evidence propagates
through its observed transitions. We achieve this by defining Bellman backups and
their fixed point, which yields behaviour-consistent state and action values.}

\paragraph{Backup operators.}
Let $\mathcal{V}_g$ denote the value functions $V:\mathcal{X}_g\to[0,1]$ with $V(z^{+})=1$ and $V(z^{-})=0$. The action backup averages discounted successor values under the empirical transition kernel, and the state backup averages these action values under the empirical behaviour policy:
\begin{equation}
\begin{aligned}
(\mathcal{B}_gV)(s,a)
&:=\beta\sum_{x\in\mathcal{X}_g}P_g(x\mid s,a)\,V(x),
&& s\in\mathcal{S}_g,\ a\in\mathcal{A}_g(s),\\[2pt]
(\mathcal{T}_gV)(s)
&:=\sum_{a\in\mathcal{A}_g(s)}\mu_g(a\mid s)\,
(\mathcal{B}_gV)(s,a),
&& s\in\mathcal{S}_g.
\end{aligned}
\label{eq:backups}
\end{equation}
The state backup keeps both boundary values fixed. Each application of $\mathcal{T}_g$ performs one round of cross-rollout recombination: evidence at successor anchors propagates back to $s$ across one additional observed edge, including evidence from rollouts that did not visit $s$. By contrast, \eqref{eq:v0} pools complete realized suffix returns without recombining continuations across rollouts. Repeated backups extend this propagation along longer observed paths.

\paragraph{The behaviour-policy fixed point.}
We define the closure of this propagation as a value assignment unchanged by further Bellman backups. Each anchor's value then equals the empirical average of its discounted successor values, yielding the fixed-point condition $\mathcal{T}_gV_g=V_g$. We express this condition as a linear system by collecting the transitions among non-terminal anchors and the one-step contribution from the success boundary into
\begin{equation}
M_g(s,s'):=
\sum_{a\in\mathcal{A}_g(s)}
\mu_g(a\mid s)P_g(s'\mid s,a),
\qquad
b_g(s):=
\beta\sum_{a\in\mathcal{A}_g(s)}
\mu_g(a\mid s)P_g(z^{+}\mid s,a),
\label{eq:matrix-form}
\end{equation}
where $s,s'\in\mathcal{S}_g$. Restricting values to these non-terminal anchors, the state backup is affine, $\mathcal{T}_gV=\beta M_gV+b_g$. The matrix $M_g$ is substochastic: its row sums are at most one because probability mass may leave to the absorbing boundaries. The fixed-point condition therefore becomes $(I-\beta M_g)V_g=b_g$, so well-posedness reduces to the invertibility of $I-\beta M_g$.

\begin{proposition}[Well-posedness]
\label{prop:wellposed}
For any group and any $\beta\in(0,1)$, $\mathcal{T}_g$ is a $\beta$-contraction on $\mathcal{V}_g$ in $\|\cdot\|_{\infty}$ and admits a unique fixed point $V_g\in\mathcal{V}_g$. Bellman iteration converges to $V_g$ from every initializer in $\mathcal{V}_g$, and $I-\beta M_g$ is invertible.
\end{proposition}
The closure is therefore independent of the initializer and has the closed form
\begin{equation}
V_g=(I-\beta M_g)^{-1}b_g,
\qquad
Q_g:=\mathcal{B}_gV_g.
\label{eq:fixed-point}
\end{equation}
In practice, we obtain $V_g$ by solving $(I-\beta M_g)V_g=b_g$ directly, without forming the inverse or iterating $\mathcal{T}_g$ (Appendix~\ref{app:wellposed}). 
With no intermediate rewards and boundary values $V(z^{+})=1$ and $V(z^{-})=0$, $V_g(s)$ gives the \emph{exact} $\beta$-discounted probability of reaching success from $s$ under $\mu_g$ on the empirical process induced by the current group.

\subsection{Finite depth and the reference direction}
\label{sec:family}
\tr{We next relate visit-local estimation to the full closure through a finite-depth family. We then characterize when feasible policy reweighting using these credits does not decrease discounted success value on the fixed empirical process.}
\paragraph{The finite-depth family.}
We initialize Bellman iteration at the visit-local estimator in \eqref{eq:v0}, with the same fixed boundary values. For integers $K\ge 1$,
\begin{equation}
V_g^{(K)}:=\mathcal{T}_g^{K}V_g^{(0)},
\qquad
Q_g^{(K)}:=\mathcal{B}_gV_g^{(K-1)},
\qquad
A_g^{(K)}:=Q_g^{(K)}-V_g^{(K)},
\label{eq:qk-vk}
\end{equation}
with $A_g^{(0)}:=Q_g^{(0)}-V_g^{(0)}$. As $K\to\infty$, the value estimates converge to $V_g$ and $Q_g$ in \eqref{eq:fixed-point}, and the limiting credit is denoted by $A_g^{(\infty)}:=Q_g-V_g$. Here $K$ counts rounds of recombination rather than steps of lookahead, since \eqref{eq:v0} already carries complete realized suffixes at $K=0$. 
All members remain centred under $\mu_g$ at each anchor, and varying $K$ changes reach while holding the empirical process and aggregation fixed. 
The $K=0$ member uses visit-local suffix averaging \citep{feng2025group}, whereas shortest-path estimators \citep{cheng2026beyond} lie outside this family because they aggregate by an extremum rather than an expectation.

On the fixed empirical process, let $V_g^\nu$ denote the $\beta$-discounted success value under a policy $\nu$ on the observed action support, and let $M_g^\nu$ be obtained from $M_g$ in \eqref{eq:matrix-form} by replacing $\mu_g$ with $\nu$. In particular, $V_g^{\mu_g}=V_g$.

\begin{proposition}[The closure credit as a reference direction]
\label{prop:alignment}
Let $\tilde{A}$ be centred under $\mu_g$ at each anchor, let $w:\mathcal{S}_g\to[0,\infty)$, and define $\xi(s,a):=\mu_g(a\mid s)w(s)\tilde{A}(s,a)$. For any $\alpha\ge0$ such that $\nu_\alpha:=\mu_g+\alpha\xi$ is a valid policy,
\begin{equation}
V_g^{\nu_\alpha}-V_g
=
\alpha\bigl(I-\beta M_g^{\nu_\alpha}\bigr)^{-1}
\Bigl(
w\odot
\bigl\langle\tilde{A},A_g^{(\infty)}\bigr\rangle_{\mu_g}
\Bigr),
\label{eq:alignment}
\end{equation}
where $\langle f,h\rangle_{\mu_g}(s):=\sum_a\mu_g(a\mid s)f(s,a)h(s,a)$ and $\odot$ denotes componentwise multiplication. For $\tilde{A}=A_g^{(\infty)}$, the inner product is the anchor-wise variance $\sigma_g(s)^2:=\sum_a\mu_g(a\mid s)[A_g^{(\infty)}(s,a)]^2\ge0$, giving $V_g^{\nu_\alpha}\succeq V_g$ componentwise.
\end{proposition}

Finite-depth credits inherit this guarantee when their deviation from the closure is sufficiently small.

\begin{corollary}[Finite-depth alignment]
\label{cor:depth}
For $K\ge1$, let $\bar{\delta}_g^{(0)}:=\mathcal{T}_gV_g^{(0)}-V_g^{(0)}$ be the empirical Bellman residual on non-terminal anchors. Then
\begin{equation}
\bigl\|A_g^{(\infty)}-A_g^{(K)}\bigr\|_{\infty}
\le
\frac{2\beta^K}{1-\beta}
\bigl\|\bar{\delta}_g^{(0)}\bigr\|_{\infty}.
\label{eq:depth-bound}
\end{equation}
Under the assumptions of Proposition~\ref{prop:alignment}, taking $\tilde{A}=A_g^{(K)}$ gives $V_g^{\nu_\alpha}\succeq V_g$ whenever the right-hand side of \eqref{eq:depth-bound} does not exceed $\sigma_g(s)$ at every anchor with $w(s)>0$ and $\sigma_g(s)>0$; anchors with $\sigma_g(s)=0$ contribute a zero inner product in \eqref{eq:alignment}.
\end{corollary}

Together, Proposition~\ref{prop:alignment} and Corollary~\ref{cor:depth} establish closure credit as a reference direction on the fixed empirical process: feasible policy reweighting along this direction does not decrease discounted success value, and sufficiently accurate finite-depth credits inherit the same guarantee. Proofs are provided in Appendix~\ref{app:proofs}, with further analysis of Bellman residuals and finite-depth errors in Appendix~\ref{app:residual}.

\subsection{Step credit and policy optimization}
\label{sec:credit}
\tr{Finally, we convert closure values into normalized step-level credit and
combine it with the trajectory-level group advantage for critic-free policy
optimization.}

\paragraph{Step-level advantage.}
For CRBC, we gate closure credit on anchors with at least two distinct observed actions and normalise it by the anchor-wise standard deviation (Figure~\ref{fig:framework}(b)):
\begin{equation}
A^{\mathrm S}_g(s,a)
:=
\frac{\mathbf{1}\{|\mathcal{A}_g(s)|\ge 2\}}
{\max\{\sigma_g(s),\,\sigma_{\min}\}}
\bigl(Q_g(s,a)-V_g(s)\bigr).
\label{eq:step-credit}
\end{equation}
Here $\sigma_g(s)$ is defined in Proposition~\ref{prop:alignment}, and $\sigma_{\min}>0$ guards nearly flat rows. Denote the non-negative prefactor by $c_g(s)$; the same proposition then applies with $w=c_g$, preserving alignment. All occurrences of the same (anchor, semantic action) pair share this credit.

\paragraph{Policy optimization.}
CRBC retains the trajectory-level advantage $A_i^{\mathrm G}$ (Figure~\ref{fig:framework}(c)) so that steps without an identifiable anchor comparison remain covered. The advantage substituted into \eqref{eq:base-objective} at step $(i,t)$ is
\begin{equation}
A^{\mathrm{final}}_{i,t}
=
w_{\mathrm G}A^{\mathrm G}_i
+
w_{\mathrm S}A^{\mathrm S}_g(s_{i,t},a_{i,t}),
\label{eq:final-advantage}
\end{equation}
with weights $w_{\mathrm G},w_{\mathrm S}\ge0$, giving the CRBC objective
\begin{equation}
\mathcal{J}_{\mathrm{CRBC}}(\theta)
=
\mathbb{E}\biggl[\frac{1}{\sum_{i}T_i}\sum_{i,t}
\min\bigl(\rho_{i,t}A^{\mathrm{final}}_{i,t},\;
\mathrm{clip}(\rho_{i,t},1\pm\epsilon)A^{\mathrm{final}}_{i,t}\bigr)\biggr]
-\lambda_{\mathrm{KL}}\mathbb{D}_{\mathrm{KL}}\bigl[\pi_{\theta}\|\pi_{\mathrm{ref}}\bigr],
\label{eq:crbc-objective}
\end{equation}
where $\sum_{i}T_i$ is the total number of environment steps in the group, and the remaining notation ($\rho_{i,t}$, $\epsilon$, $\lambda_{\mathrm{KL}}$) is that of \eqref{eq:base-objective}. Only the advantage differs from \eqref{eq:base-objective}, so CRBC introduces no parametric critic and no additional rollouts. The advantage and the empirical process behind it receive no gradient.

\begin{table}[t]
\centering
\caption{Performance comparison on ALFWorld and WebShop. For ALFWorld, we report the average success rate (\%) for each subtask and the overall success rate. For WebShop, we report the average task score and average success rate (\%). Most results are averaged over three random seeds. The best results are highlighted in \textbf{bold}.}
\label{tab:main-results}
\vskip 0.08in
\scriptsize
\setlength{\tabcolsep}{1.5pt}
\begin{tabular}{@{}ll|ccccccc|cc@{}}
\toprule
& & \multicolumn{7}{c|}{\textbf{ALFWorld}}
& \multicolumn{2}{c@{}}{\textbf{WebShop}} \\
\cmidrule(lr){3-9}\cmidrule(l){10-11}
\textbf{Type} & \textbf{Method}
& Pick & Clean & Cool & Look & Heat & Pick2 & All
& Score & Succ. \\
\midrule

\multicolumn{11}{@{}l}{\textit{\textbf{Closed-Source Models}}} \\
Prompting & GPT-4o
& 75.30 & 60.80 & 31.20 & 56.70 & 21.60 & 49.80 & 48.00
& 31.80 & 23.70 \\
Prompting & Gemini-2.5-Pro
& 92.80 & 63.30 & 62.10 & 69.00 & 26.60 & 58.70 & 60.30
& 42.50 & 35.90 \\

\midrule
\multicolumn{11}{@{}l}{\textit{\textbf{Qwen2.5-1.5B-Instruct}}} \\

Prompting & Qwen2.5
& 5.90 & 5.50 & 3.30 & 9.70 & 4.20 & 0.00 & 4.10
& 23.10 & 5.20 \\

Prompting & ReAct
& 17.40 & 20.50 & 15.70 & 6.20 & 7.70 & 2.00 & 12.80
& 40.10 & 11.30 \\

Prompting & Reflexion
& 35.30 & 22.20 & 21.70 & 13.60 & 19.40 & 3.70 & 21.80
& 55.80 & 21.90 \\

RL Training & PPO
& 64.80\sd{3.50}
& 40.50\sd{6.90}
& 57.10\sd{4.90}
& 60.60\sd{6.60}
& 46.40\sd{4.00}
& 47.40\sd{1.90}
& 54.30\sd{3.10}
& 73.80\sd{3.00}
& 51.50\sd{2.90} \\

RL Training & RLOO
& 88.30\sd{3.00}
& 52.80\sd{8.60}
& 71.00\sd{5.90}
& 62.80\sd{8.70}
& 66.40\sd{5.50}
& 56.90\sd{4.70}
& 69.70\sd{2.50}
& 73.90\sd{5.60}
& 52.10\sd{6.70} \\

RL Training & GRPO
& 85.27\sd{1.40}
& 65.79\sd{2.63}
& 88.08\sd{8.08}
& 58.33\sd{0.00}
& 78.57\sd{7.14}
& 75.00\sd{5.00}
& 77.73\sd{1.95}
& 83.33\sd{3.52}
& 71.09\sd{0.78} \\

RL Training & GiGPO
& 96.67\sd{4.71}
& 84.21\sd{4.30}
& 88.21\sd{3.02}
& 72.22\sd{3.93}
& 95.24\sd{3.89}
& 88.33\sd{2.36}
& 89.32\sd{1.95}
& 88.93\sd{1.55}
& 75.78\sd{2.21} \\

RL Training & GraphGPO
& 95.63\sd{3.09}
& 84.21\sd{4.30}
& 88.31\sd{0.22}
& 88.89\sd{3.93}
& 92.06\sd{5.94}
& 88.33\sd{2.36}
& 90.10\sd{1.84}
& 89.60\sd{1.46}
& 79.69\sd{1.28} \\

RL Training & HGPO
& 94.30\sd{2.90}
& 69.10\sd{9.20}
& \textbf{95.90}\sd{2.80}
& \textbf{97.40}\sd{3.60}
& 92.00\sd{2.30}
& 82.30\sd{3.20}
& 90.50\sd{0.40}
& 85.50\sd{0.50}
& 70.50\sd{1.70} \\

\rowcolor{gray!15}
\textbf{RL Training} & \textbf{CRBC (Ours)}
& \textbf{100.00}\sd{0.00}
& \textbf{92.98}\sd{6.56}
& 92.15\sd{3.33}
& 91.67\sd{0.00}
& \textbf{96.83}\sd{4.49}
& \textbf{100.00}\sd{0.00}
& \textbf{96.09}\sd{0.64}
& \textbf{90.11}\sd{1.24}
& \textbf{81.51}\sd{0.74} \\

\midrule
\multicolumn{11}{@{}l}{\textit{\textbf{Qwen2.5-7B-Instruct}}} \\

Prompting & Qwen2.5
& 33.40 & 21.60 & 19.30 & 6.90 & 2.80 & 3.20 & 14.80
& 26.40 & 7.80 \\

Prompting & ReAct
& 48.50 & 35.40 & 34.30 & 13.20 & 18.20 & 17.60 & 31.20
& 46.20 & 19.50 \\

Prompting & Reflexion
& 62.00 & 41.60 & 44.90 & 30.90 & 36.30 & 23.80 & 42.70
& 58.10 & 28.80 \\

RL Training & PPO
& 92.30\sd{4.00}
& 64.00\sd{8.40}
& 92.50\sd{2.40}
& 89.50\sd{7.00}
& 80.30\sd{2.00}
& 68.80\sd{8.30}
& 80.40\sd{2.70}
& 81.40\sd{3.10}
& 68.70\sd{5.10} \\

RL Training & RLOO
& 87.60\sd{4.30}
& 78.20\sd{8.30}
& 87.30\sd{5.80}
& 81.30\sd{7.60}
& 71.90\sd{5.20}
& 48.90\sd{8.40}
& 75.50\sd{4.60}
& 80.30\sd{3.20}
& 65.70\sd{4.00} \\

RL Training & GRPO
& 88.98\sd{5.30}
& 91.98\sd{4.43}
& 77.89\sd{4.58}
& 78.57\sd{0.00}
& \textbf{90.74}\sd{5.24}
& 71.43\sd{3.89}
& 83.33\sd{2.05}
& 84.31\sd{1.27}
& 75.00\sd{2.78} \\

RL Training & GiGPO
& \textbf{100.00}\sd{0.00}
& 92.11\sd{7.89}
& 80.77\sd{7.69}
& 95.83\sd{4.17}
& 90.48\sd{0.00}
& 90.00\sd{5.00}
& 91.41\sd{2.34}
& 86.60\sd{1.17}
& 76.56\sd{1.56} \\

RL Training & GraphGPO
& 97.85\sd{3.04}
& 89.47\sd{7.44}
& 88.31\sd{0.22}
& 94.44\sd{3.93}
& 82.54\sd{8.09}
& 86.67\sd{6.24}
& 90.10\sd{3.51}
& 87.24\sd{1.57}
& 78.13\sd{0.00} \\

\rowcolor{gray!15}
\textbf{RL Training} & \textbf{CRBC (Ours)}
& \textbf{100.00}\sd{0.00}
& \textbf{100.00}\sd{0.00}
& \textbf{94.23}\sd{5.77}
& \textbf{100.00}\sd{0.00}
& 90.48\sd{0.00}
& \textbf{97.50}\sd{2.50}
& \textbf{96.88}\sd{1.56}
& \textbf{90.43}\sd{1.25}
& \textbf{81.51}\sd{3.51} \\

\bottomrule
\end{tabular}

\vspace{-10pt}
\end{table}

\section{Experiments}
\label{sec:experiments}

Our experiments answer three questions: (1) does cross-rollout Bellman closure improve LLM-agent training over trajectory-level and step-level baselines; (2) how does it behave over the course of training; and (3) how do its two design choices (combining step- and trajectory-level credit, and taking the closure depth to $K\!\to\!\infty$) affect performance.

\subsection{Experiment Setup}



\paragraph{Benchmarks and baselines.}
We evaluate CRBC on text-based \textbf{ALFWorld} \citep{shridhar2021alfworld} and \textbf{WebShop} \citep{yao2022webshop} for household interaction and web shopping, respectively, and on visual $6\times6$ \textbf{Sokoban} \citep{SchraderSokoban2018}. Baselines include closed-source models GPT-4o \citep{achiam2023gpt} and Gemini-2.5-Pro \citep{geminiteam2025geminifamilyhighlycapable}; prompting baselines Qwen2.5 \citep{qwen2025qwen25technicalreport}, ReAct \citep{yao2023reactsynergizingreasoningacting}, and Reflexion \citep{shinn2023reflexionlanguageagentsverbal}; and RL methods PPO \citep{schulman2017proximal}, RLOO \citep{ahmadian2024basicsrevisitingreinforcestyle}, GRPO \citep{deepseek-math}, GiGPO \citep{feng2025group}, GraphGPO \citep{cheng2026beyond}, and HGPO \citep{he2026hierarchyofgroupspolicyoptimizationlonghorizon}. For Sokoban, group-based RL baselines share the same VLM backbone. Benchmark details and baseline descriptions, including implementation provenance, appear in Appendices~\ref{app:environment-details} and~\ref{app:comparing-methods}, respectively.

\paragraph{Implementation details.}
We use Qwen2.5-1.5B-Instruct and Qwen2.5-7B-Instruct \citep{qwen2025qwen25technicalreport} for ALFWorld and WebShop, and Qwen2.5-VL-3B-Instruct \citep{bai2025qwen25vltechnicalreport} for Sokoban. Within each benchmark and model scale, reproduced RL methods share the same data, rollout, and optimization protocol, differing only in the advantage estimator and its method-specific settings. Each update uses $G=8$ rollouts per group, with 16 groups for ALFWorld/WebShop and 32 for Sokoban, and episodes capped at 50, 15, and 15 environment steps, respectively. Rollout and validation temperatures are 1.0 and 0.4, the actor learning rate is $10^{-6}$, and the KL coefficient is 0.01; we train for 150 updates and validate every five. All local runs use one node of eight 80\,GB NVIDIA A100 GPUs. Unless stated otherwise, CRBC uses $\beta=0.98$, full closure ($K\!\to\!\infty$), and $(w_{\mathrm G},w_{\mathrm S})=(1,5)$ without auxiliary predictors. Full configurations appear in Appendix~\ref{app:training-details}.

\subsection{Experimental Results}
\label{sec:exp-results}

\textbf{Performance on agentic benchmarks.} Table~\ref{tab:main-results} reports the comparison between CRBC and the baselines on ALFWorld and WebShop, and Table~\ref{tab:sokoban-results} reports the results on Sokoban. Across both model sizes and all three environments, CRBC consistently outperforms every baseline in overall success. On ALFWorld with Qwen2.5-1.5B-Instruct, CRBC reaches an overall success rate of $96.09\%$, improving over GRPO by $18.36$ points and over the strongest baseline (HGPO, $90.50\%$) by $5.59$ points; it improves over GRPO on every subtask, ranks first on four of the six, and solves Pick and Pick2 perfectly. On WebShop, CRBC attains the best task score ($90.11$) and success rate ($81.51\%$). The advantage persists at the 7B scale, where CRBC achieves $96.88\%$ on ALFWorld and $81.51\%$ success on WebShop, surpassing all baselines on both overall metrics. On the vision-language Sokoban benchmark (Table~\ref{tab:sokoban-results}), CRBC reaches $82.81\%$ success, exceeding GraphGPO by $5.21$ points, GiGPO by $6.25$ points, and GRPO by a large margin. These results show that recovering high-fidelity step-level credit through cross-rollout Bellman closure translates into consistent end-task gains across embodied, web, and visual multi-turn settings.

\begin{figure*}[t]
\centering
\includegraphics[
    width=\textwidth
]{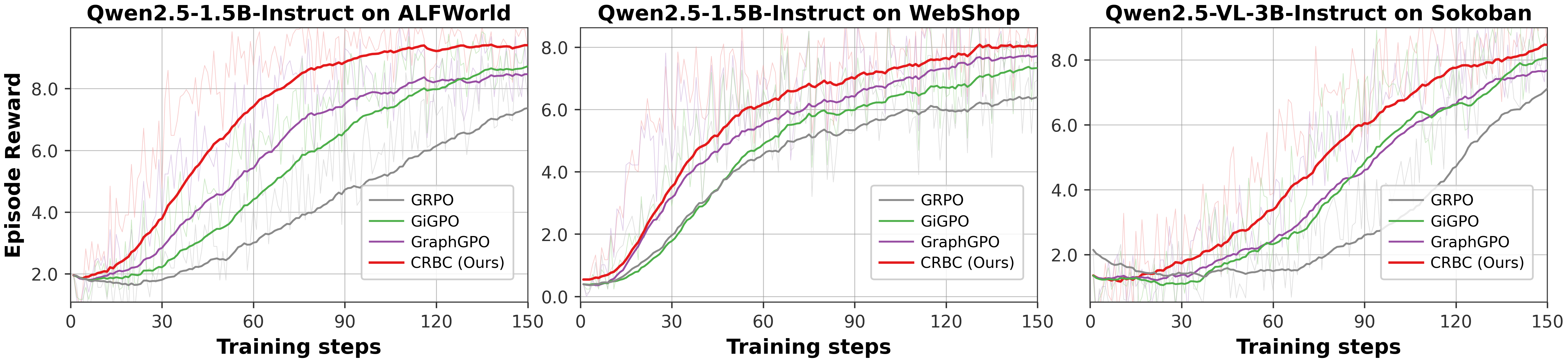}

\caption{
Training reward curves for CRBC (Ours, red), GraphGPO (blue), GiGPO (green),
and GRPO (gray) on ALFWorld, WebShop, and Sokoban. Light and dark curves
denote per-update and EMA-smoothed rewards ($\alpha=0.95$), respectively.
Qwen2.5-1.5B-Instruct is used for ALFWorld/WebShop and
Qwen2.5-VL-3B-Instruct for Sokoban; panel scales are independent.
}
\label{fig:training-reward-curves}
\vspace{-5pt}
\end{figure*}

\textbf{Training dynamics.}
Figure~\ref{fig:training-reward-curves} shows mean trajectory reward for CRBC, GraphGPO, GiGPO, and GRPO, with validation success rates in Figure~\ref{fig:appendix-training-success-curves} (appendix). Across the three environments, step-level methods improve faster and attain higher rewards than trajectory-level GRPO. CRBC rises fastest during early and middle training and achieves the highest final reward; on ALFWorld with Qwen2.5-1.5B-Instruct, it matches GraphGPO's final reward in roughly half the updates. Similar validation trends show improved held-out success alongside the reward gains, supporting the learning-efficiency benefit of cross-rollout Bellman closure.


\begin{table}[t]
\centering

\begin{minipage}[t]{0.49\linewidth}
\caption{Closure-depth ablation on ALFWorld with Qwen2.5-1.5B-Instruct. Only $K$ varies; all other settings are fixed. Best values are in \textbf{bold}.}
\label{tab:depth-ablation}
\end{minipage}\hfill
\begin{minipage}[t]{0.49\linewidth}
\caption{Test performance on $6\times6$ Sokoban with Qwen2.5-VL-3B-Instruct. Results are averaged over three seeds; best values are in \textbf{bold}.}
\label{tab:sokoban-results}
\end{minipage}

\par\vspace{0.4em}

\begin{minipage}[t]{0.48\linewidth}
\vspace{0pt}
\centering
\scriptsize
\setlength{\tabcolsep}{2pt}
\renewcommand{\arraystretch}{1.25}

\begin{tabularx}{\linewidth}{@{}*{4}{>{\centering\arraybackslash}X}@{}}
\toprule
\textbf{Depth $K$}
& \textbf{Step-credit}
& \textbf{Score}
& \textbf{Success (\%)} \\
\midrule

$0$
& $A^{(0)}$
& 7.00\sd{0.27}
& 92.97\sd{0.00} \\

$1$
& $A^{(1)}$
& 7.05\sd{0.53}
& 92.97\sd{1.56} \\

$2$
& $A^{(2)}$
& 7.16\sd{1.20}
& 93.75\sd{3.12} \\

$3$
& $A^{(3)}$
& 7.21\sd{0.13}
& 93.75\sd{0.78} \\

\rowcolor{gray!15}
$\infty$
& $A^{(\infty)}$
& \textbf{8.01}\sd{0.27}
& \textbf{96.09}\sd{0.64} \\

\bottomrule
\end{tabularx}
\end{minipage}\hfill
\begin{minipage}[t]{0.48\linewidth}
\vspace{0pt}
\centering
\scriptsize
\setlength{\tabcolsep}{3pt}
\renewcommand{\arraystretch}{1.15}

\begin{tabularx}{\linewidth}{@{}ll*{2}{>{\centering\arraybackslash}X}@{}}
\toprule
\textbf{Type}
& \textbf{Method}
& \textbf{Score}
& \textbf{Success (\%)} \\
\midrule

Prompting
& Qwen2.5-VL$^\ast$
& ---
& 11.70 \\

\midrule

RL Training
& GRPO$^\dagger$
& 1.81\sd{2.50}
& 46.48\sd{3.64} \\

RL Training
& GiGPO
& 4.00\sd{1.12}
& 76.56\sd{5.47} \\

RL Training
& GraphGPO
& 4.13\sd{0.34}
& 77.60\sd{2.88} \\

\rowcolor{gray!15}
\textbf{RL Training}
& \textbf{CRBC (Ours)}
& \textbf{5.06}\sd{0.54}
& \textbf{82.81}\sd{2.30} \\

\bottomrule
\end{tabularx}
\end{minipage}

\end{table}
\begin{table*}[t]
\centering
\caption{Credit and weight ablations on ALFWorld with Qwen2.5-1.5B-Instruct. We report final validation success rates (\%) for six subtasks and the overall success rate (All) at step 150, averaged over three random seeds. Best values are in \textbf{bold}.}
\label{tab:alfworld-credit-ablation}
\vskip 0.08in

\scriptsize
\setlength{\tabcolsep}{2.3pt}
\renewcommand{\arraystretch}{1.20}

\begin{tabular*}{\textwidth}{@{\extracolsep{\fill}}ccl|ccccccc@{}}
\toprule
$\boldsymbol{w}_{\mathrm G}$
& $\boldsymbol{w}_{\mathrm S}$
& \textbf{Credit assignment}
& \textbf{Pick}
& \textbf{Clean}
& \textbf{Cool}
& \textbf{Look}
& \textbf{Heat}
& \textbf{Pick2}
& \textbf{All} \\
\midrule

$0$ & $5$ & Step-level only
& 98.89\sd{1.57}
& \textbf{98.25}\sd{2.48}
& 93.54\sd{1.74}
& \textbf{94.44}\sd{3.93}
& 88.89\sd{2.24}
& 96.67\sd{2.36}
& 95.31\sd{1.69} \\

$1$ & $0$ & Trajectory-level only
& 84.62\sd{1.46}
& 64.91\sd{2.48}
& 86.92\sd{6.79}
& 58.33\sd{0.00}
& 77.78\sd{5.94}
& 73.33\sd{4.71}
& 76.56\sd{2.30} \\

\midrule

$1$ & $1$ & Step + trajectory
& 98.89\sd{1.57}
& 87.72\sd{2.48}
& 94.82\sd{1.78}
& \textbf{94.44}\sd{3.93}
& 88.89\sd{2.24}
& 93.33\sd{2.36}
& 93.75\sd{1.69} \\

$1$ & $2.5$ & Step + trajectory
& 98.89\sd{1.57}
& 92.98\sd{4.96}
& \textbf{97.38}\sd{1.85}
& 86.11\sd{3.93}
& 92.06\sd{2.24}
& 93.33\sd{2.36}
& 94.53\sd{1.91} \\

\rowcolor{gray!15}[6pt][6pt]
$1$ & $5$ & \textbf{Step + trajectory (Ours)}
& \textbf{100.00}\sd{0.00}
& 92.98\sd{6.56}
& 92.15\sd{3.33}
& 91.67\sd{0.00}
& \textbf{96.83}\sd{4.49}
& \textbf{100.00}\sd{0.00}
& \textbf{96.09}\sd{0.64} \\

$1$ & $10$ & Step + trajectory
& 98.89\sd{1.57}
& 92.98\sd{2.48}
& 94.87\sd{3.63}
& 91.67\sd{0.00}
& 88.89\sd{2.24}
& 98.33\sd{2.36}
& 94.79\sd{1.95} \\

\bottomrule
\end{tabular*}

\vspace{1mm}
\end{table*}

\subsection{Ablation Study}
\label{sec:exp-ablation}

\textbf{Closure depth.}
Table~\ref{tab:depth-ablation} varies only the closure depth $K$, keeping the aggregation rule, anchor map $\phi$, and other settings fixed. Figure~\ref{fig:appendix-training-abtion} shows that reward, validation success, and validation score generally improve faster as $K$ increases, with $K=3$ and full closure showing the strongest early trends. Final success rates are $92.97\%$ for $K=0,1$, $93.75\%$ for $K=2,3$, and $96.09\%$ for full closure, a $3.12$-point gain over the visit-local estimator; validation score also rises from $7.00$ at $K=0$ to $8.01$ under full closure. Figure~\ref{fig:closure-coverage} provides a structural explanation using rollouts from an early-training checkpoint: the proportion of eligible visits with access to all eight rollouts increases from $12.9\%$ at $K=0$ to $61.5\%$ under full closure, while the paired heatmap shows expanded coverage for $63.0\%$ of eligible visits. These results support the benefit of deeper Bellman propagation and illustrate how closure makes additional cross-rollout evidence accessible beyond the current anchor.

\textbf{Credit assignment and weight sensitivity.}
Table~\ref{tab:alfworld-credit-ablation} evaluates the trajectory-level and step-level weights $w_{\mathrm G}$ and $w_{\mathrm S}$. Trajectory-level credit alone $(1,0)$ achieves $76.56\%$ overall success, whereas step-level closure credit alone $(0,5)$ reaches $95.31\%$. With $w_{\mathrm G}=1$, increasing $w_{\mathrm S}$ from $1$ to $2.5$ and $5$ raises success from $93.75\%$ to $94.53\%$ and $96.09\%$; a further increase to $10$ reduces it to $94.79\%$. These results suggest that closure credit provides the main gains, complemented by appropriately weighted trajectory-level credit. The default $(1,5)$ achieves the highest overall success, exceeding step-level credit alone by $0.78$ points.

\begin{wrapfigure}{r}{0.52\textwidth}
\vspace{-0.5\baselineskip}
\centering
\includegraphics[width=\linewidth]{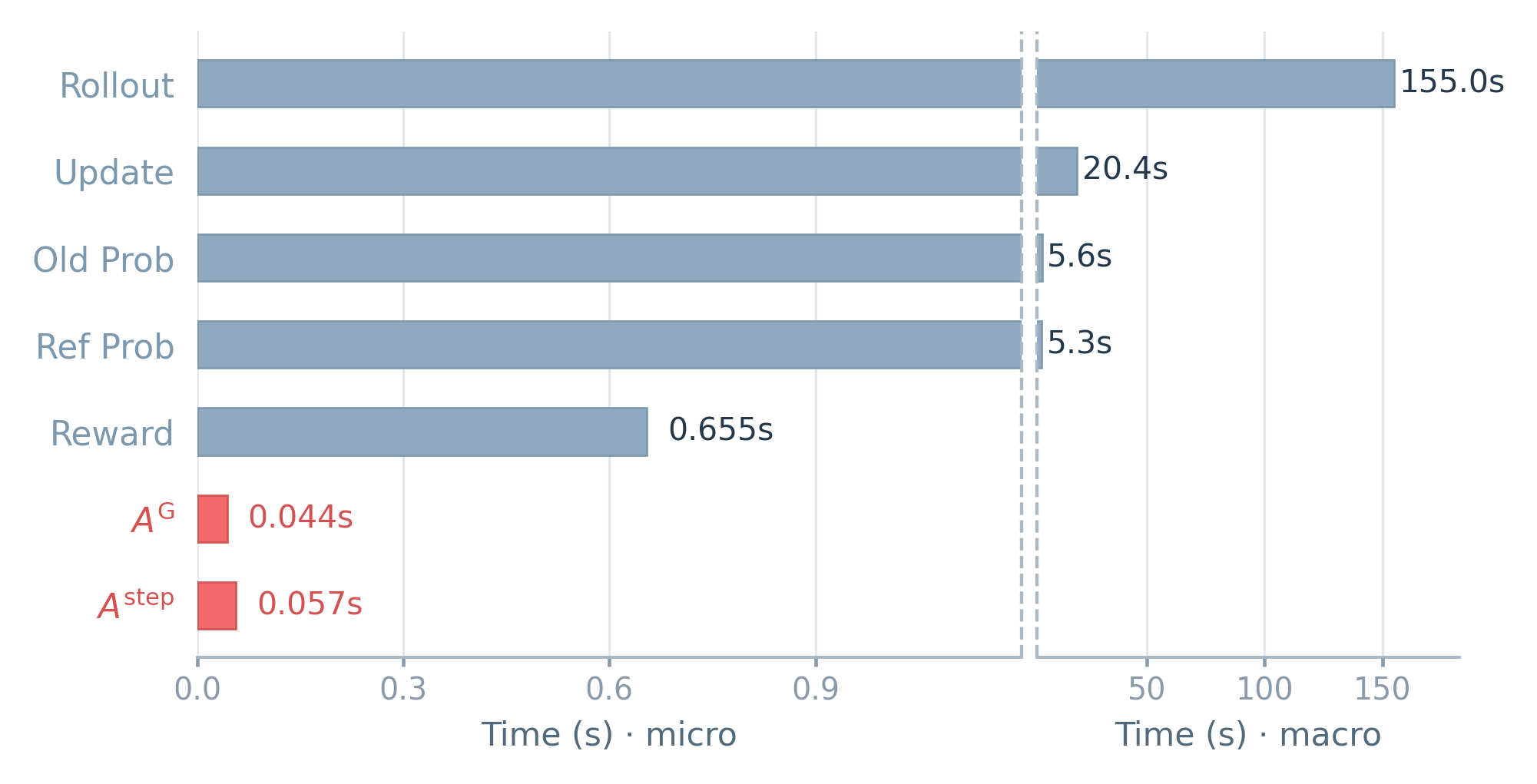}
\caption{Per-update runtime breakdown of CRBC on ALFWorld with Qwen2.5-1.5B-Instruct. Blue bars denote shared stages, while red bars denote trajectory-level and step-level credit construction.}
\label{fig:runtime-crbc}
\vspace{-0.5\baselineskip}
\end{wrapfigure}

\textbf{Computational overhead.}
Figure~\ref{fig:runtime-crbc} presents the per-update runtime breakdown of CRBC on ALFWorld with Qwen2.5-1.5B-Instruct, averaged over 120 non-validation updates from a profiling run. An update takes $187.46$\,s on average, dominated by rollout ($154.99$\,s) and the policy update ($20.35$\,s). The trajectory-level group advantage $A^{\mathrm G}$ and the instrumented CRBC step-credit core $A^{\mathrm S}$ take only $0.044$\,s and $0.057$\,s, respectively, corresponding to $0.02\%$ and $0.03\%$ of the total update time. Including tensor preparation and advantage assembly, the complete advantage block takes $0.330$\,s, or $0.18\%$ of the update. Since CRBC reuses the same rollout and actor-update pipeline without introducing additional model rollouts or a critic network, its credit construction remains negligible relative to the main training stages.
\section{Conclusion}
\label{sec:conclusion}

We presented CRBC, which evaluates each rollout group's empirical process at its behavior-policy Bellman fixed point to derive critic-free $Q-V$ credit with global cross-rollout propagation and behavior-consistent aggregation. Feasible policy reweighting along this credit does not decrease discounted success value on the fixed process. Combined with trajectory-level advantages, the normalized credit improves performance and learning efficiency on ALFWorld, WebShop, and Sokoban without additional rollouts, supported by depth and credit ablations.





\bibliography{iclr2027_conference}
\bibliographystyle{iclr2027_conference}

\newpage
\appendix

\section{Algorithm}

\begin{algorithm}[htbp]
\caption{CRBC with full Bellman closure}
\label{alg:CRBC}
\begin{algorithmic}[1]
\REQUIRE Initial policy $\pi_\theta$, fixed reference policy $\pi_{\mathrm{ref}}$, task distribution $p$, anchor map $\phi$, discount $\beta\in(0,1)$, advantage weights $w_{\mathrm G},w_{\mathrm S}$, clipping parameter $\epsilon$, KL penalty $\lambda_{\mathrm{KL}}$, group size $G$, normalization floor $\sigma_{\min}>0$

\FOR{each training iteration}
    \STATE Set $\theta_{\mathrm{old}}\leftarrow\theta$ and sample a batch of tasks from $p$

    \FOR{each task $g$ in the batch}
        \STATE {\small\color{gray}// Group rollout}
        \STATE Initialise $G$ environments from the same task condition
        \STATE Collect complete rollouts using $y_{i,t}\sim\pi_{\theta_{\mathrm{old}}}(\cdot\mid h_{i,t})$ and $a_{i,t}=\mathrm{Exec}(y_{i,t})$; retain contexts, responses, and terminal outcomes $R_i$

        \STATE {\small\color{gray}// Empirical process and Bellman closure}
        \STATE Map $s_{i,t}=\phi(h_{i,t})$ and construct $\mathcal{D}_g$ with successors $x_{i,t}$ and absorbing boundaries as in Section~\ref{sec:prelim}
        \STATE Build $(P_g,\mu_g)$ from transition counts using \eqref{eq:empirical-model}
        \STATE Form $M_g,b_g$ using \eqref{eq:matrix-form} and solve $(I-\beta M_g)V_g=b_g$
        \STATE Set $V_g(z^+)=1$ and $V_g(z^-)=0$, then compute $Q_g=\mathcal{B}_gV_g$

        \STATE {\small\color{gray}// Credit construction and combination}
        \STATE Compute trajectory-level group-relative advantages $A_i^{\mathrm G}$ using the underlying optimizer
        \STATE Compute $A_g^{(\infty)}(s,a)=Q_g(s,a)-V_g(s)$ and its standard deviation $\sigma_g(s)$ under $\mu_g(\cdot\mid s)$ at each anchor
        \STATE Compute the gated and normalized step-level credit $A_g^{\mathrm S}$ using \eqref{eq:step-credit}
        \STATE Set $A_{i,t}^{\mathrm{final}}=w_{\mathrm G}A_i^{\mathrm G}+w_{\mathrm S}A_g^{\mathrm S}(s_{i,t},a_{i,t})$ for every step in the group
    \ENDFOR

    \STATE {\small\color{gray}// Policy update}
    \STATE Hold advantages and graph statistics fixed, and update $\theta$ using the objective $\mathcal{J}_{\mathrm{CRBC}}(\theta)$ in \eqref{eq:crbc-objective}
\ENDFOR
\end{algorithmic}
\end{algorithm}

Algorithm~\ref{alg:CRBC} summarizes the full-closure training procedure. At each iteration, CRBC constructs an empirical process for each rollout group, solves its behaviour-policy Bellman fixed point, and combines the resulting normalized step credit with trajectory-level group-relative advantages for policy optimization.

\section{Proofs}
\label{app:proofs}

All value functions, including $V^{(0)}_g$, have fixed boundary values $V(z^{+})=1$ and $V(z^{-})=0$. Matrix expressions act on their restrictions to $\mathcal{S}_g$, where $\mathcal{T}_gV=\beta M_gV+b_g$ and $\|M_g\|_{\infty}\le1$. For a state vector $u$ and a policy $\nu$ on the observed action support, define
\begin{equation}
\begin{gathered}
(P^{S}_gu)(s,a):=\sum_{s'\in\mathcal{S}_g}P_g(s'\mid s,a)\,u(s'),
\qquad
(Lu)(s,a):=u(s),\\[4pt]
M^{\nu}_g(s,s'):=\sum_{a\in\mathcal{A}_g(s)}\nu(a\mid s)\,P_g(s'\mid s,a).
\end{gathered}
\label{eq:app-operators}
\end{equation}
These operators are non-expansive in the sup norm. Let $\mathcal{T}^{\nu}_g$ be the state backup with $\mu_g$ replaced by $\nu$, and $V^{\nu}_g$ its fixed point. Since boundary differences vanish, for $U,W\in\mathcal{V}_g$,
\begin{equation}
(\mathcal{B}_gU)(s,a)-(\mathcal{B}_gW)(s,a)
=\beta\bigl(P^{S}_g(U-W)\bigr)(s,a).
\label{eq:app-backup-diff}
\end{equation}

\subsection{Proof of Proposition~\ref{prop:wellposed}}
\label{app:wellposed}

\paragraph{Invariance.}
For $V\in\mathcal{V}_g$ and $s\in\mathcal{S}_g$,
\begin{equation*}
0\le(\mathcal{T}_gV)(s)
=\beta\sum_a\mu_g(a\mid s)\sum_x P_g(x\mid s,a)V(x)
\le\beta<1.
\end{equation*}
Together with the fixed boundary values, this shows that $\mathcal{T}_g$ maps $\mathcal{V}_g$ into itself.

\paragraph{Contraction and conclusion.}
For $U,V\in\mathcal{V}_g$, substochasticity gives
\begin{equation*}
\|\mathcal{T}_gU-\mathcal{T}_gV\|_{\infty}
=\beta\|M_g(U-V)\|_{\infty}
\le\beta\|U-V\|_{\infty}.
\end{equation*}
The space $\mathcal{V}_g$ is closed in the finite-dimensional sup-norm space and hence complete. The Banach fixed point theorem therefore gives a unique fixed point, and Bellman iteration converges to it from every initializer with error at most $\beta^K$ times the initial error. Moreover, $V_g(s)\in[0,\beta]$ for $s\in\mathcal{S}_g$. Since $\|\beta M_g\|_{\infty}\le\beta<1$, the inverse $(I-\beta M_g)^{-1}=\sum_{k\ge0}(\beta M_g)^k$ exists, is entrywise non-negative, and has norm at most $(1-\beta)^{-1}$. The fixed-point equation then yields \eqref{eq:fixed-point}. \qed

\subsection{Proof of Proposition~\ref{prop:alignment}}
\label{app:alignment}

\paragraph{Feasible direction.}
Centring gives $\sum_a\xi(s,a)=w(s)\sum_a\mu_g(a\mid s)\tilde{A}(s,a)=0$, so $\nu_\alpha$ sums to one at each anchor. Non-negativity holds for $0\le\alpha\le\alpha_\xi$, where $\alpha_\xi:=\min\{\mu_g(a\mid s)/|\xi(s,a)|:\xi(s,a)<0\}$, with $\alpha_\xi=\infty$ when no negative entries exist.

\paragraph{Value difference.}
Using the fixed-point equation and centring of $\tilde{A}$,
\begin{equation}
\begin{aligned}
\bigl(\mathcal{T}^{\nu_\alpha}_gV_g-V_g\bigr)(s)
&=\sum_a\bigl[\nu_\alpha(a\mid s)-\mu_g(a\mid s)\bigr]Q_g(s,a)\\
&=\alpha\,w(s)\langle\tilde{A},A^{(\infty)}_g\rangle_{\mu_g}(s).
\end{aligned}
\label{eq:app-one-step}
\end{equation}
Subtracting the two Bellman equations gives $\bigl(I-\beta M^{\nu_\alpha}_g\bigr)\bigl(V^{\nu_\alpha}_g-V_g\bigr)=\mathcal{T}^{\nu_\alpha}_gV_g-V_g$, establishing \eqref{eq:alignment}. The resolvent uses the perturbed policy $\nu_\alpha$ and is entrywise non-negative by its Neumann-series representation.

\paragraph{Self-alignment.}
For $\tilde{A}=A^{(\infty)}_g$,
\begin{equation*}
\bigl\langle A^{(\infty)}_g,A^{(\infty)}_g\bigr\rangle_{\mu_g}(s)
=\operatorname{Var}_{a\sim\mu_g(\cdot\mid s)}
\bigl[Q_g(s,a)\bigr]
=\sigma_g(s)^2\ge0.
\end{equation*}
The forcing and resolvent are therefore non-negative, yielding $V^{\nu_\alpha}_g\succeq V_g$. Improvement is strict at $s$ when $\alpha>0$ and an anchor with $w>0$ and $\sigma_g>0$ is reachable from $s$ under $\nu_\alpha$, including a path of length zero. \qed

\subsection{Residual decomposition and finite-depth transport}
\label{app:residual}

\paragraph{Residual definitions.}
Define the action-level residual and its state-wise average:
\begin{equation}
\begin{aligned}
\delta^{(0)}_g(s,a)
&:=(\mathcal{B}_gV^{(0)}_g)(s,a)-Q^{(0)}_g(s,a)
=Q^{(1)}_g(s,a)-Q^{(0)}_g(s,a),\\[3pt]
\bar{\delta}^{(0)}_g(s)
&:=\sum_a\mu_g(a\mid s)\,\delta^{(0)}_g(s,a)
=(\mathcal{T}_gV^{(0)}_g)(s)-V^{(0)}_g(s).
\end{aligned}
\label{eq:app-residuals}
\end{equation}
Action-level residuals can cancel in this average, so $\bar{\delta}^{(0)}_g\equiv0$ need not imply $\delta^{(0)}_g\equiv0$.

\paragraph{Provenance.}
For an observed transition $(s,a,x)$, define the suffix evidence arriving from $(s,a)$ by
\begin{equation}
V^{(0)}_{g\leftarrow(s,a)}(x):=
\frac{1}{N_g(s,a,x)}
\sum_{\substack{(i,t):\,s_{i,t}=s,\ a_{i,t}=a,\\ x_{i,t}=x}}
\beta^{T_i-t}R_i.
\label{eq:app-incoming}
\end{equation}
Set this quantity to zero for unobserved successors, whose transition probabilities are zero. Factoring $\beta^{T_i-t+1}R_i=\beta(\beta^{T_i-t}R_i)$ and grouping occurrences by successor gives
\begin{equation*}
Q^{(0)}_g(s,a)
=\beta\sum_{x\in\mathcal{X}_g}P_g(x\mid s,a)\,
V^{(0)}_{g\leftarrow(s,a)}(x).
\end{equation*}
Subtracting this from the action backup yields
\begin{equation}
\delta^{(0)}_g(s,a)
=\beta\sum_{x\in\mathcal{X}_g}P_g(x\mid s,a)
\Bigl[V^{(0)}_g(x)-V^{(0)}_{g\leftarrow(s,a)}(x)\Bigr].
\label{eq:app-provenance}
\end{equation}
Boundary contributions vanish. For $x\in\mathcal{S}_g$, $V^{(0)}_g(x)=N_g(x)^{-1}\sum_{(i,t)\in\mathcal{I}_g(x)}G_{i,t}$ averages all occurrences at $x$, whereas the incoming average includes only those reached from $(s,a)$. Thus the residual vanishes if every observed non-terminal successor has no initial occurrences and receives transitions only from $(s,a)$, or if the weighted discrepancies cancel. Acyclicity alone does not suffice, since different state-action pairs may share a successor.

\paragraph{Transport.}
Since $(I-\beta M_g)(V_g-V^{(0)}_g)=\bar{\delta}^{(0)}_g$, set $d_g:=(I-\beta M_g)^{-1}\bar{\delta}^{(0)}_g=V_g-V^{(0)}_g$. The affine Bellman recurrence gives, by induction,
\begin{equation}
V_g-V^{(K)}_g
=(\beta M_g)^Kd_g,
\qquad K\ge0.
\label{eq:app-finite-value}
\end{equation}
For $K\ge1$, \eqref{eq:app-backup-diff} then gives
\begin{equation}
\begin{aligned}
Q_g-Q^{(K)}_g
&=\beta P^{S}_g(\beta M_g)^{K-1}d_g,\\
A^{(\infty)}_g-A^{(K)}_g
&=\Bigl[\beta P^{S}_g(\beta M_g)^{K-1}
-L(\beta M_g)^K\Bigr]d_g.
\end{aligned}
\label{eq:app-finite-qa}
\end{equation}
At $K=0$, adding and subtracting $\mathcal{B}_gV^{(0)}_g$ instead yields
\begin{equation}
\begin{aligned}
Q_g-Q^{(0)}_g
&=\delta^{(0)}_g+\beta P^{S}_gd_g,\\
A^{(\infty)}_g-A^{(0)}_g
&=\delta^{(0)}_g+\bigl(\beta P^{S}_g-L\bigr)d_g.
\end{aligned}
\label{eq:app-transport}
\end{equation}
Consequently, the $K=0$ deviation cannot be controlled solely by $\|\bar{\delta}^{(0)}_g\|_\infty$: when $\bar{\delta}^{(0)}_g\equiv0$, we have $d_g=0$, but $A^{(\infty)}_g-A^{(0)}_g=\delta^{(0)}_g$ may remain nonzero. Considering the sequences from $K=0$ onward, the state-value sequence is constant if and only if $\bar{\delta}^{(0)}_g\equiv0$, whereas the paired sequence $(V^{(K)}_g,Q^{(K)}_g)$ is constant if and only if $\delta^{(0)}_g\equiv0$.

\subsection{Proof of Corollary~\ref{cor:depth}}
\label{app:tolerance}

\paragraph{Depth bound.}
For $K\ge1$, non-expansiveness of the operators in \eqref{eq:app-finite-qa} and the bound on $(I-\beta M_g)^{-1}$ give
\begin{equation*}
\bigl\|A_g^{(\infty)}-A_g^{(K)}\bigr\|_\infty
\le2\beta^K\|d_g\|_\infty
\le\frac{2\beta^K}{1-\beta}
\bigl\|\bar{\delta}_g^{(0)}\bigr\|_\infty
=:E_K,
\end{equation*}
establishing \eqref{eq:depth-bound}. The restriction $K\ge1$ follows from the additional action-level residual in \eqref{eq:app-transport}.

\paragraph{Alignment and value improvement.}
For a credit $\tilde{A}$ centred under $\mu_g$, write $e:=A_g^{(\infty)}-\tilde{A}$ and use the weighted norm $\|e(s,\cdot)\|_{\mu_g}:=\bigl(\sum_a\mu_g(a\mid s)e(s,a)^2\bigr)^{1/2}$. Cauchy--Schwarz gives, anchor-wise,
\begin{equation}
\bigl\langle\tilde{A},A_g^{(\infty)}\bigr\rangle_{\mu_g}
=\sigma_g^2-
\bigl\langle e,A_g^{(\infty)}\bigr\rangle_{\mu_g}
\ge\sigma_g
\bigl(\sigma_g-\|e\|_{\mu_g}\bigr).
\label{eq:envelope}
\end{equation}
Taking $\tilde{A}=A_g^{(K)}$, which is centred under $\mu_g$, gives $\|e(s,\cdot)\|_{\mu_g}\le\|e\|_\infty\le E_K$. Hence the inner product is non-negative whenever $E_K\le\sigma_g(s)$. If $\sigma_g(s)=0$, then $A_g^{(\infty)}(s,\cdot)=0$ on the observed support, so the inner product is zero regardless of $E_K$. At anchors with $w(s)=0$, the forcing term in \eqref{eq:alignment} is also zero. Thus, if $E_K\le\sigma_g(s)$ at every anchor with $w(s)>0$ and $\sigma_g(s)>0$, the forcing is componentwise non-negative. For any feasible reweighting from Proposition~\ref{prop:alignment}, the non-negative resolvent in \eqref{eq:alignment} then yields $V_g^{\nu_\alpha}\succeq V_g$.

\paragraph{Existence of a finite depth.}
Let $\mathcal{S}_{g,+}(w):=\{s\in\mathcal{S}_g:w(s)>0,\sigma_g(s)>0\}$. If this set is non-empty, its finiteness gives $\min_{s\in\mathcal{S}_{g,+}(w)}\sigma_g(s)>0$. Since $E_K\to0$, the sufficient condition holds for all sufficiently large finite $K$. If the set is empty, all forcing terms are zero and the condition holds vacuously. Failure to satisfy this sufficient condition does not imply misalignment. The decay of $E_K$ improves the certified lower bound $\sigma_g(\sigma_g-E_K)$, but does not imply monotone alignment, preserved action rankings, or a particular training-curve shape. \qed

\section{Experiment Details}
\label{app:details}

\subsection{Comparing Methods}
\label{app:comparing-methods}

We compare CRBC with the following baselines.
\begin{itemize}
    \item \textbf{GPT-4o}: a closed-source, general-purpose LLM used as a reference for multi-turn agent capability~\citep{achiam2023gpt}.
    \item \textbf{Gemini-2.5-Pro}: a second closed-source LLM reference point with strong general-purpose reasoning ability~\citep{geminiteam2025geminifamilyhighlycapable}.
    \item \textbf{ReAct}: an in-context prompting agent that interleaves reasoning and acting~\citep{yao2023reactsynergizingreasoningacting}.
    \item \textbf{Reflexion}: a prompting agent that uses verbal reflection and iterative self-improvement without parameter updates~\citep{shinn2023reflexionlanguageagentsverbal}.
    \item \textbf{PPO}: the standard clipped policy-gradient method with a learned value critic and GAE~\citep{schulman2017proximal}.
    \item \textbf{RLOO}: a critic-free, trajectory-level leave-one-out baseline~\citep{ahmadian2024basicsrevisitingreinforcestyle}.
    \item \textbf{GRPO}: a critic-free group objective that normalizes one outcome advantage over the rollouts of a task~\citep{deepseek-math}.
    \item \textbf{GiGPO}: a step-level method that compares occurrences at repeated anchor states~\citep{feng2025group}.
    \item \textbf{GraphGPO}: a graph-based method that scores transitions by a shortest-path/extremal readout~\citep{cheng2026beyond}.
    \item \textbf{HGPO}: a hierarchical grouping method that refines step comparisons using historical-context consistency~\citep{he2026hierarchyofgroupspolicyoptimizationlonghorizon}.
    \item \textbf{CRBC (ours)}: the predictor-free, full-closure setting described in Appendix~\ref{app:training-details}.
\end{itemize}

\begin{table}[t]
\centering
\caption{Environment-level training settings used in our reproduced experiments. Settings are shared across methods within each model scale and environment.}
\label{tab:app-env-settings}

\scriptsize
\setlength{\tabcolsep}{4pt}
\renewcommand{\arraystretch}{1.15}

\begin{tabular*}{\linewidth}{@{\extracolsep{\fill}}lccc@{}}
\toprule
\textbf{Setting}
& \textbf{ALFWorld}
& \textbf{WebShop}
& \textbf{Sokoban} \\
\midrule
Maximum environment steps & 50 & 15 & 15 \\
Maximum prompt tokens & 2048 & 5120 & 1024 \\
Maximum response tokens & 512 & 512 & 512 \\
Training groups per update & 16 & 16 & 32 \\
Rollouts per group & 8 & 8 & 8 \\
Validation trajectories & 128 & 128 & 128 \\
Base model
& Qwen2.5-1.5B/7B
& Qwen2.5-1.5B/7B~\citep{qwen2025qwen25technicalreport}
& Qwen2.5-VL-3B~\citep{bai2025qwen25vltechnicalreport} \\
\bottomrule
\end{tabular*}
\end{table}

\subsection{Environment Details}
\label{app:environment-details}

\paragraph{ALFWorld.}
{ALFWorld}~\citep{shridhar2021alfworld} is a text-based embodied environment in which an agent must complete a household task through a long-horizon sequence of admissible commands. We use the six task families reported by the environment evaluator: Pick, Clean, Cool, Look, Heat, and Pick2. The training wrapper is \texttt{AlfredTWEnv}; it augments the textual observation with the current location, holding/item status, item-transition history, and the sorted admissible-command list. Success and failure are terminal outcomes, and a trajectory is truncated after at most 50 environment steps. Each validation pass evaluates 128 in-distribution trajectories selected by the environment seed.

\paragraph{WebShop.}
{WebShop}~\citep{yao2022webshop} is a text-based interactive shopping environment in which the agent searches, navigates, and selects an item that matches a natural-language request. We use the repository's text-rich, small-catalogue configuration (\texttt{observation\_mode=text\_rich} and \texttt{use\_small=True}, using the repository's \texttt{items\_shuffle\_1000.json}/\texttt{items\_ins\_v2\_1000.json} files) with two retained interaction records. The \texttt{WebAgentTextEnv} wrapper limits a trajectory to 15 environment steps; the validation goals use indices 0--499 and training starts at index 500 in the launcher. We report both the raw task score and the binary task success rate.

\paragraph{Sokoban.}
{Sokoban}~\citep{SchraderSokoban2018} tests transfer to a visual interactive setting. We use the $6\times6$ one-box configuration with \texttt{Qwen2.5-VL-3B-Instruct}~\citep{bai2025qwen25vltechnicalreport}; the checkpoint is pinned to the revision used by the releazed launcher. The environment returns an RGB observation and the agent chooses among the four box-pushing directions. Trajectories are limited to 15 environment steps. We report the final raw task score (``\texttt{text\_score}'') and the percentage of solved boards (``\texttt{success\_rate}'').

\paragraph{Anchor construction.}
Anchors are matched by exact equality of their environment-specific representations within each task group. ALFWorld uses the current textual observation augmented with location, held-item status, item-transition records, and sorted admissible commands; WebShop uses the formatted current-page observation; Sokoban uses the current RGB observation. The recent interaction records appended to policy prompts are excluded from the matching key.

\subsection{Details of Training}
\label{app:training-details}

\paragraph{Shared protocol.}
For each update, text experiments sample 16 task groups with $G=8$ rollouts per group, giving 128 training rollouts. Sokoban uses 32 groups with the same group size. All settings use 128 validation trajectories, validate before training and every five updates, and train for 150 updates. The rollout temperature is 1.0 and the validation temperature is 0.4 with sampling enabled. The actor learning rate is $10^{-6}$, the reference-policy KL-loss coefficient is 0.01 (the low-variance KL form), and gradient checkpointing is enabled. For ALFWorld and WebShop, a successful trajectory receives reward 10 and failure receives 0. An invalid-action penalty of $0.1$ is applied before advantage computation. PPO alone uses a learned critic with learning rate $10^{-5}$; group-based methods do not train an auxiliary critic. Environment and data-loader seeds are set to the same value for each run, and means and standard deviations are computed only over completed independent seeds.

\begin{figure}[t]
    \centering
    \includegraphics[width=\linewidth]{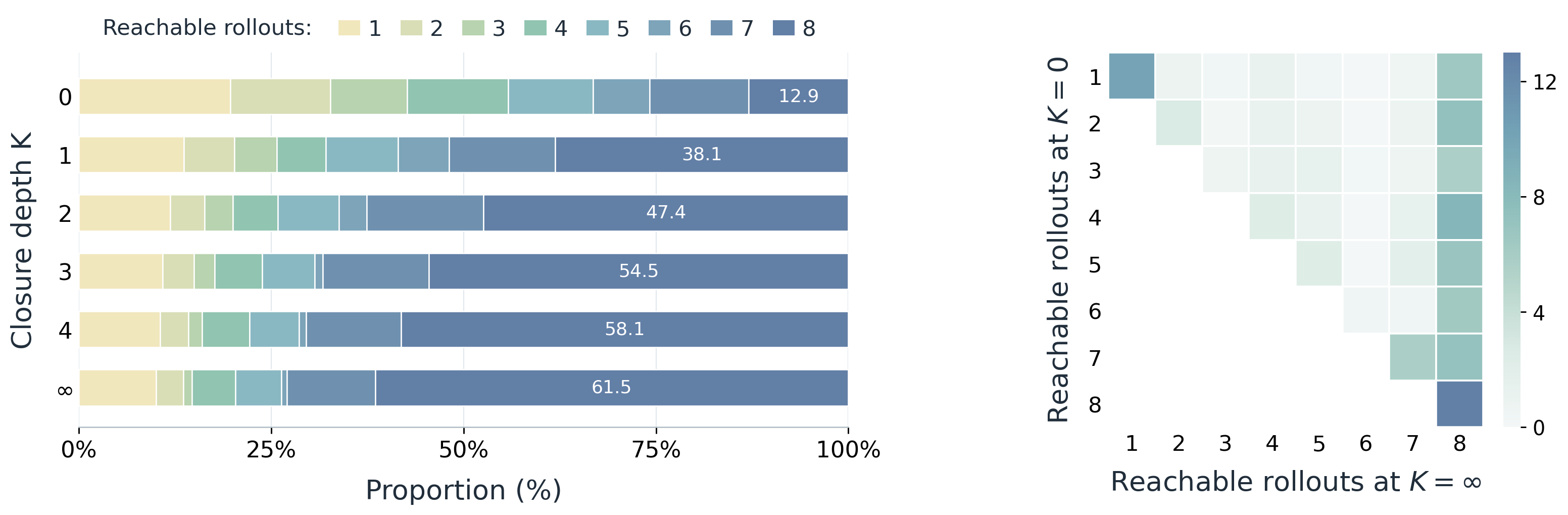}
    \caption{Cross-rollout evidence coverage measured on rollouts from an early-training checkpoint on ALFWorld. Left: distribution of the number of reachable rollouts at each closure depth $K$. Right: paired coverage at $K=0$ and full closure, with color indicating the percentage of all eligible visits.}
    \label{fig:closure-coverage}
\end{figure}

\paragraph{Resource and batching settings.}
The batching choices are adjusted only to fit the model and environment; within each comparison they are identical across methods. For ALFWorld, the 1.5B and 7B runs use tensor parallel sizes 1 and 4, actor micro-batches 32 and 4 per GPU, and PPO mini-batches 256 and 64, respectively. For WebShop, the corresponding tensor parallel sizes are 2 and 4, actor micro-batches are 8 and 4, and the PPO mini-batch is 64 for both scales. Sokoban uses tensor parallel size 2, actor micro-batch 8, and PPO mini-batch 64. The rollout and reference log-probability micro-batches are 32 (ALFWorld 1.5B), 16 (ALFWorld 7B and WebShop), and 16 (Sokoban). All reproduced experiments are conducted on a single node equipped with eight NVIDIA A100 GPUs (80 GB each); tensor parallelism and micro-batches are adjusted within this fixed budget as specified above.

\paragraph{Method-specific settings.}
GraphGPO uses its default shortest-path readout with graph discounts 0.10, 0.20, and 0.80 on ALFWorld, WebShop, and Sokoban, respectively, and step/trajectory weights $1/1$. GiGPO uses a return discount of $0.95$ and step/trajectory weights $1/1$ (the final Sokoban launcher uses its \texttt{mean\_norm} variant). GRPO uses the standard group-normalized trajectory advantage, while PPO and RLOO use GAE and a trajectory-level leave-one-out baseline, respectively. Unless otherwise stated, CRBC uses the full-closure configuration with $K=\infty$, $\beta=0.98$, $w_{\mathrm G}=1$, $w_{\mathrm S}=5$, and a normalization floor $\sigma_{\min}=0.1$. The empirical process is formed from the observed rollout transitions, and the resulting $Q-V$ signal is normalized per anchor before being combined with the trajectory-level advantage. All other training settings follow the common protocol above.

\begin{table}[t]
\centering
\caption{Bellman-discount ablation on ALFWorld with Qwen2.5-1.5B-Instruct. Only $\beta$ varies; all other CRBC settings are fixed. Results are averaged over three random seeds; best values are in \textbf{bold}.}
\label{tab:beta-ablation}
\vskip 0.08in

\scriptsize
\setlength{\tabcolsep}{2.4pt}
\renewcommand{\arraystretch}{1.25}

\begin{tabular}{@{}c|ccccccc@{}}
\toprule
\textbf{Bellman discount $\beta$}
& Pick
& Clean
& Cool
& Look
& Heat
& Pick2
& All \\
\midrule

$0.95$
& 98.89\sd{1.57}
& \textbf{98.25}\sd{2.48}
& \textbf{94.82}\sd{1.78}
& 91.67\sd{0.00}
& 88.89\sd{2.24}
& 98.33\sd{2.36}
& 95.57\sd{0.37}
\\

\rowcolor{gray!15}
$0.98$
& \textbf{100.00}\sd{0.00}
& 92.98\sd{6.56}
& 92.15\sd{3.33}
& 91.67\sd{0.00}
& \textbf{96.83}\sd{4.49}
& \textbf{100.00}\sd{0.00}
& \textbf{96.09}\sd{0.64}
\\

$0.99$
& \textbf{100.00}\sd{0.00}
& 86.84\sd{2.63}
& 94.23\sd{5.77}
& \textbf{95.83}\sd{4.17}
& 92.86\sd{2.38}
& \textbf{100.00}\sd{0.00}
& 95.31\sd{1.56}
\\

$1.00$
& \textbf{100.00}\sd{0.00}
& 89.47\sd{10.53}
& 92.23\sd{3.77}
& 83.33\sd{8.33}
& 95.24\sd{4.76}
& 92.50\sd{2.50}
& 93.36\sd{0.39}
\\

\bottomrule
\end{tabular}

\vspace{1mm}
\end{table}

\begin{table*}[t]
\centering
\caption{Rollout group size comparison on ALFWorld with Qwen2.5-1.5B-Instruct at step 150. We report validation success rates (\%) for six subtasks and overall (All), together with task score. Best reported values within each group size are in \textbf{bold}.}
\label{tab:group-size-ablation}
\vskip 0.08in

\scriptsize
\setlength{\tabcolsep}{2.3pt}
\renewcommand{\arraystretch}{1.20}

\begin{tabularx}{\textwidth}{cl|*{8}{>{\centering\arraybackslash}X}}
\toprule
\textbf{Group size $G$}
& \textbf{Method}
& \textbf{Pick}
& \textbf{Clean}
& \textbf{Cool}
& \textbf{Look}
& \textbf{Heat}
& \textbf{Pick2}
& \textbf{All}
& \textbf{Score} \\
\midrule

$4$ & GRPO
& 60.00
& 36.84
& 46.15
& 50.00
& 38.10
& 50.00
& 47.66
& 2.27 \\

$4$ & GiGPO
& 90.00
& 63.16
& \textbf{88.46}
& 66.67
& 80.95
& 75.00
& 79.69
& 5.34 \\

$4$ & GraphGPO
& 86.67
& 68.42
& 84.62
& 58.33
& 76.19
& 75.00
& 77.34
& 4.38 \\

$4$ & \textbf{CRBC (Ours)}
& \textbf{100.00}
& \textbf{94.74}
& \textbf{88.46}
& \textbf{91.67}
& \textbf{100.00}
& \textbf{80.00}
& \textbf{92.97}
& \textbf{7.12} \\

\midrule

$16$ & GRPO
& 93.33
& 73.68
& 80.77
& 58.33
& 85.71
& 80.00
& 81.25
& 5.08 \\

$16$ & GiGPO
& \textbf{100.00}
& 89.47
& 88.46
& 83.33
& \textbf{100.00}
& 90.00
& 92.97
& 7.01 \\

$16$ & GraphGPO
& \textbf{100.00}
& 78.95
& 88.46
& 66.67
& 90.48
& 85.00
& 87.50
& 5.30 \\

$16$ & \textbf{CRBC (Ours)}
& \textbf{100.00}
& \textbf{94.74}
& \textbf{96.15}
& \textbf{100.00}
& 90.48
& \textbf{100.00}
& \textbf{96.88}
& \textbf{8.34} \\

\bottomrule
\end{tabularx}

\vspace{1mm}
\end{table*}

\section{Additional Experiments}
\label{app:additional-experiments}


\subsection{Bellman Discount}
\label{app:beta-ablation}

Table~\ref{tab:beta-ablation} evaluates the Bellman discount $\beta$ on ALFWorld with Qwen2.5-1.5B-Instruct, keeping all other CRBC settings fixed. All results are averaged over three random seeds, with the default $\beta=0.98$ using the main CRBC result. Across the tested discounted settings $\beta\in\{0.95,0.98,0.99\}$, overall success remains between $95.31\%$ and $96.09\%$, with $\beta=0.98$ achieving the highest observed mean. Removing discounting ($\beta=1.00$) reduces overall success to $93.36\%$, a decrease of $2.73$ percentage points relative to the default.

Subtask preferences vary: $\beta=0.95$ performs best on Clean and Cool, $\beta=0.99$ on Look, and $\beta=0.98$ on Heat. All configurations use full closure, so $\beta$ controls how strongly distant evidence is attenuated along the merged graph. These results suggest that mild discounting is beneficial in this setting and support $\beta=0.98$ as a practical default, with limited sensitivity among the tested discounted values.

\subsection{Closure Depth and Evidence Coverage}
\label{app:depth-analysis}

Figure~\ref{fig:closure-coverage} examines cross-rollout evidence coverage using rollouts from an early-training checkpoint on ALFWorld. We consider visits to anchors with at least two distinct observed actions. For each eligible visit, coverage counts the distinct rollouts whose evidence is structurally accessible at depth $K$ within its eight-rollout group, without weighting their contributions. The left panel shows the coverage distribution: increasing $K$ allows evidence from shared successors to propagate back through the merged graph, shifting visits toward broader coverage. The proportion of eligible visits with access to all eight rollouts increases from $12.9\%$ at $K=0$ to $61.5\%$ under full closure.

\begin{figure*}[t]
\centering
\includegraphics[
width=\textwidth
]{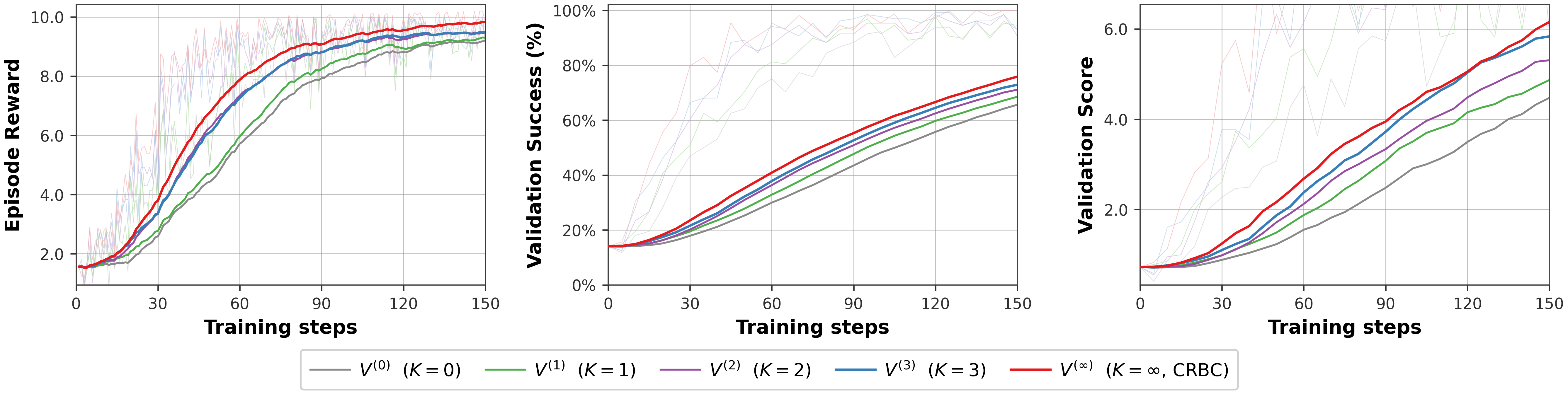}

\caption{Training dynamics for the closure-depth ablation on ALFWorld with
Qwen2.5-1.5B-Instruct. Only the closure depth $K$ varies; all other settings
are fixed. From left to right, we show episode reward, validation success
rate, and validation score. Light and dark curves denote raw and
EMA-smoothed trajectories ($\alpha=0.95$); deeper closure gives stronger
late-training performance.}

\label{fig:appendix-training-abtion}
\end{figure*}

The right panel pairs coverage before and after closure for the same visits, with rows representing $K=0$, columns representing full closure, and color intensity indicating the percentage of all eligible visits. Diagonal cells indicate unchanged coverage; cells with larger column values indicate expansion. Overall, coverage increases for $63.0\%$ of eligible visits, including those whose expanded coverage remains below eight rollouts. This structural view complements the final performance comparison in Table~\ref{tab:depth-ablation} and the corresponding training curves in Figure~\ref{fig:appendix-training-abtion}, illustrating how closure makes additional cross-rollout evidence available.

\begin{figure*}[t]
\centering
\includegraphics[
width=\textwidth
]{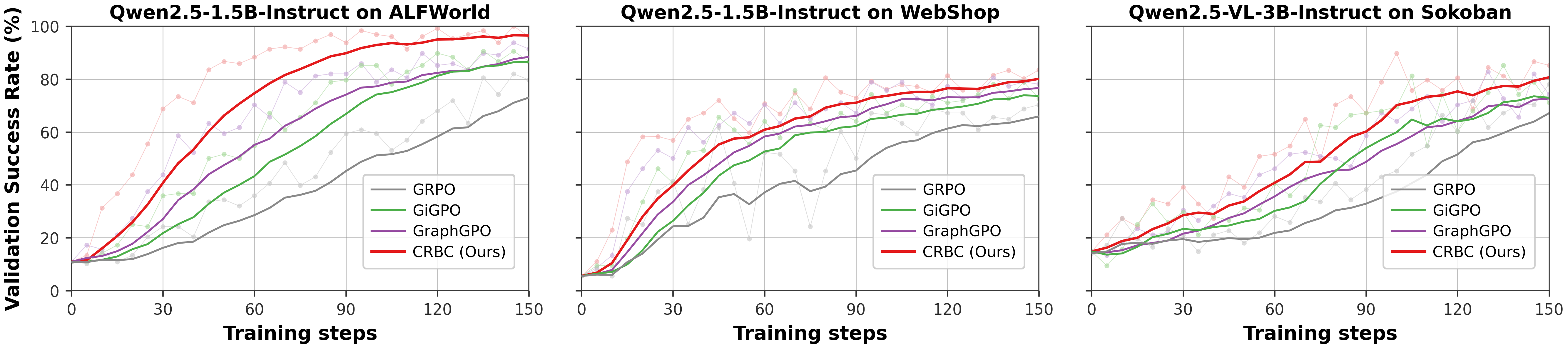}

\caption{
Validation success-rate curves for CRBC (Ours, red), GraphGPO (purple),
GiGPO (green), and GRPO (gray) on ALFWorld, WebShop, and Sokoban.
Light curves show validation measurements recorded every five training
updates, while dark curves show EMA-smoothed trends
($\alpha=0.95$ per update). Qwen2.5-1.5B-Instruct is used for
ALFWorld and WebShop, and Qwen2.5-VL-3B-Instruct for Sokoban.
}

\label{fig:appendix-training-success-curves}
\end{figure*}
\subsection{Validation Success}

Figure~\ref{fig:appendix-training-success-curves} reports validation success rates on ALFWorld, WebShop, and Sokoban, measured every five training updates. CRBC generally reaches high success rates earlier and achieves stronger final performance than the compared baselines. These trends complement the training reward curves in Figure~\ref{fig:training-reward-curves}, showing that the observed learning-efficiency gains are also reflected in task success on the validation sets.

\subsection{Rollout Group Size}
\label{app:group-size-ablation}

Table~\ref{tab:group-size-ablation} compares group-based RL methods with rollout group sizes $G\in\{4,16\}$ on ALFWorld using Qwen2.5-1.5B-Instruct. At $G=4$, CRBC achieves $92.97\%$ success, compared with $47.66\%$ for GRPO, $79.69\%$ for GiGPO, and $77.34\%$ for GraphGPO. At $G=16$, CRBC reaches $96.88\%$, versus $92.97\%$ for GiGPO and $87.50\%$ for GraphGPO. CRBC therefore improves over the strongest baseline by $13.28$ and $3.91$ percentage points, respectively, while also obtaining the highest task scores of $7.12$ and $8.34$. These comparisons show that CRBC retains an advantage at both tested group sizes.

\definecolor{promptframe}{HTML}{7F8B96}
\definecolor{prompttitle}{HTML}{173B5E}
\definecolor{promptback}{HTML}{FAFAF8}
\definecolor{promptink}{HTML}{FFFFFF}
\definecolor{slottext}{HTML}{285A86}
\definecolor{tagtext}{HTML}{183B5B}

\newcommand{\promptslot}[1]{\textcolor{slottext}{\ttfamily\scriptsize\{#1\}}}
\newcommand{\prompttag}[1]{\textcolor{tagtext}{\ttfamily\scriptsize #1}}

\tcbset{promptbox/.style={enhanced,boxrule=0.65pt,arc=2.4mm,left=3.2mm,right=3.2mm,top=2.2mm,bottom=2.2mm,colback=promptback,colframe=promptframe,colbacktitle=prompttitle,coltitle=promptink,fonttitle=\bfseries\footnotesize,before skip=0pt,after skip=0pt}}

\newtcolorbox{alfworldprompt}{promptbox,title={ALFWorld\enspace|\enspace embodied action prompt}}
\newtcolorbox{webshopprompt}{promptbox,title={WebShop\enspace|\enspace shopping action prompt}}
\newtcolorbox{sokobanprompt}{promptbox,title={Sokoban\enspace|\enspace visual action prompt}}

\begin{figure*}[t]
\centering
\setlength{\emergencystretch}{2em}
\begin{minipage}{\textwidth}
\begin{alfworldprompt}
\raggedright\small
You are an expert agent operating in the ALFRED embodied environment. Your task is to: \promptslot{task\_description}. Prior to this step, you have already taken \promptslot{step\_count} step(s). Below are the most recent \promptslot{history\_length} observations and the corresponding actions you took: \promptslot{action\_history}. You are now at step \promptslot{current\_step} and your current observation is: \promptslot{current\_observation}. Your admissible actions for the current situation are: [\promptslot{admissible\_actions}].

Now it is your turn to take an action. First reason step by step about the current situation. This reasoning process \textbf{MUST} be enclosed within \prompttag{<think>...</think>}. Once you have finished your reasoning, choose an admissible action for the current step and present it within \prompttag{<action>...</action>}.
\end{alfworldprompt}
\vspace{0.6mm}
\centering\scriptsize\textbf{(a)} ALFWorld prompt
\end{minipage}
\par\vspace{2mm}
\begin{minipage}{\textwidth}
\begin{webshopprompt}
\raggedright\small
You are an expert autonomous agent operating in the WebShop e-commerce environment. Your task is to: \promptslot{task\_description}. Prior to this step, you have already taken \promptslot{step\_count} step(s). Below are the most recent \promptslot{history\_length} observations and the corresponding actions you took: \promptslot{action\_history}. You are now at step \promptslot{current\_step} and your current observation is: \promptslot{current\_observation}. Your admissible actions for the current situation are: [\promptslot{available\_actions}].

Now it is your turn to take one action for the current step. First reason step by step about the current situation, then carefully choose which admissible action best advances the shopping goal. This reasoning process \textbf{MUST} be enclosed within \prompttag{<think>...</think>}. Once you have finished your reasoning, choose an admissible action for the current step and present it within \prompttag{<action>...</action>}.
\end{webshopprompt}
\vspace{0.6mm}
\centering\scriptsize\textbf{(b)} WebShop prompt
\end{minipage}
\par\vspace{2mm}
\begin{minipage}{\textwidth}
\begin{sokobanprompt}
\raggedright\small
You are an expert agent operating in the Sokoban environment. Your goal is to push all the boxes onto the target spots. Once all boxes are on the targets, you win!

\smallskip
\textbf{Rules}\par
You can only push boxes. You can't pull them, so plan ahead to avoid getting stuck. You can't walk through or push boxes into walls. To avoid traps, do not push boxes into corners or against walls where they can't be moved again.

\smallskip
\textbf{Visual Elements in the Image}\par
\begin{minipage}[c]{0.74\linewidth}
\raggedright\small
Character: A small, green alien-like figure with two antennae and black eyes. It represents you.\par
Box: A yellow crate marked with an orange ``X'' across its front. It is the box you need to push.\par
Target: A black tile outlined in red, with a small red diamond shape in the center. It marks the destination where a box should be pushed.
\end{minipage}\hfill
\begin{minipage}[c]{0.22\linewidth}
\centering
\includegraphics[width=0.90\linewidth]{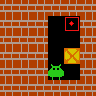}\par
{\scriptsize Example RGB input}
\end{minipage}

\smallskip
\textbf{Current Step}\par
Your current observation is shown in the image: \prompttag{<image>}\par
Your admissible actions are [``up'', ``down'', ``left'', ``right''].

\smallskip
Now it's your turn to make a move (choose ONE action only for the current step). You should first reason step-by-step about the current situation---observe the positions of boxes and targets, plan a path to push a box toward a target, and avoid traps like corners or walls. This reasoning process \textbf{MUST} be enclosed within \prompttag{<think> </think>} tags. Once you've finished your reasoning, you should choose an admissible action for current step and present it within \prompttag{<action> </action>} tags.
\end{sokobanprompt}
\vspace{0.6mm}
\centering\scriptsize\textbf{(c)} Sokoban prompt
\end{minipage}
\caption{Prompt templates for ALFWorld, WebShop, and Sokoban. Coloured placeholders denote runtime inputs. Text environments retain up to two observation--action records; Sokoban uses the current RGB observation without textual interaction history. Panel (c) includes an example RGB input.}
\label{fig:prompt-templates}
\end{figure*}

\section{Prompt Templates}
\label{app:prompts}

Figure~\ref{fig:prompt-templates} shows the prompt templates for ALFWorld, WebShop, and Sokoban. For the text environments, prompts retain at most the two most recent observation--action records. For Sokoban, the prompt contains the current RGB image, task rules, and four movement directions, without textual interaction history. In all three environments, the agent is instructed to place its reasoning inside \texttt{<think>...</think>} and its selected action inside \texttt{<action>...</action>}. Within each environment, all reproduced RL methods share the same prompt templates and memory protocol.


\end{document}